\documentclass[10pt,journal,compsoc]{IEEEtran}

\ifCLASSOPTIONcompsoc
  \usepackage[nocompress]{cite}
\else
  \usepackage{cite}
\fi
\ifCLASSINFOpdf
\else
\fi
\makeatletter
\let\NAT@parse\undefined
\makeatother
\usepackage{hyperref}  

\usepackage{amsmath}
\usepackage{amsfonts}
\usepackage{amssymb}
\usepackage{amsthm,amsmath}
\usepackage{mathrsfs}
\usepackage{indentfirst}
\usepackage{multirow}
\usepackage{color}
\usepackage{setspace}
\usepackage{makecell}
\usepackage{graphicx}

\newcommand{\blue}[1]{{\color{black} #1 }}
\newcommand{\red}[1]{{\color{black} #1 }}
\newcommand{\spara}[1]{\smallskip\noindent\textbf{#1}.}

\newtheorem{definition}{Definition}

\begin{document}
%
\title{Disentangled Representation Learning}
%
%
%

\author{Xin~Wang,~\IEEEmembership{Member,~IEEE},
        Hong~Chen, Si'ao~Tang, Zihao Wu
        and~Wenwu~Zhu,~\IEEEmembership{Fellow,~IEEE}
\IEEEcompsocitemizethanks{\IEEEcompsocthanksitem Xin Wang, Hong Chen, Si'ao~Tang, Zihao Wu and Wenwu Zhu are with the Department of Computer Science and Technology, Beijing
National Research Center for Information Science and Technology, Tsinghua University, Beijing, China. 
E-mail: \{xin\_wang,wwzhu\}@tsinghua.edu.cn, \{h-chen20,tsa22,wuzh22\}@mails.tsinghua.edu.cn. Corresponding Authors: Xin Wang, Wenwu Zhu.
\IEEEcompsocthanksitem This work was supported by the National Key Research and Development Program of China No.2023YFF1205001, National Natural Science Foundation of China (No. 62222209, 62250008, 62102222), Beijing National Research Center for Information Science and Technology under Grant No. BNR2023RC01003, BNR2023TD03006, and Beijing Key Lab of Networked Multimedia. }
}

%
%

\markboth{IEEE Transactions on Pattern Analysis and Machine Intelligence, June~2024}%
{Shell \MakeLowercase{\textit{et al.}}: Bare Demo of IEEEtran.cls for Computer Society Journals}
%



\IEEEtitleabstractindextext{%
\begin{abstract}
\red{
Disentangled Representation Learning (DRL) aims to learn a model capable of identifying and disentangling the underlying factors hidden in the observable data in representation form.
The process of separating underlying factors of variation into variables with semantic meaning benefits in learning explainable representations of data, 
which imitates the meaningful understanding process of humans when observing an object or relation. 
As a general learning strategy, DRL has demonstrated its power in improving the model explainability, controlability, robustness, as well as generalization capacity in a wide range of scenarios such as computer vision, natural language processing, and data mining. In this article, we comprehensively investigate DRL from various aspects including motivations, definitions, methodologies, evaluations, applications, and model designs. We first present two well-recognized definitions, i.e., Intuitive Definition and Group Theory Definition for disentangled representation learning. We further categorize the methodologies for DRL into four groups from the following perspectives, the model type, representation structure, supervision signal, and independence assumption. 
We also analyze principles to design different DRL models that may benefit different tasks in practical applications. Finally, we point out challenges in DRL as well as potential research directions deserving future investigations. We believe this work may provide insights for promoting the DRL research in the community.
}
\end{abstract}

}

\maketitle

\IEEEdisplaynontitleabstractindextext

%
\IEEEpeerreviewmaketitle

\IEEEraisesectionheading{\section{Introduction}\label{sec:introduction}}

%
%
%
%
When humans observe an object, we seek to understand the various properties of this object (e.g., shape, size and color etc.) with certain prior knowledge. 
However, existing end-to-end black-box deep learning models take a shortcut strategy through directly learning representations of the object to fit the data distribution and discrimination criteria~\cite{geirhos2020shortcut}, failing to extract the hidden attributes carried in representations with human-like generalization ability. 
To fill this gap, an important representation learning paradigm, \textit{Disentangled Representation Learning} (DRL) is proposed~\cite{bengio2013representation} and has attracted an increasing amount of attention in the research community.

DRL is a learning paradigm where machine learning models are designed to obtain representations capable of identifying and disentangling the underlying factors hidden in the observed data. 
DRL always benefits in learning explainable representations of the observed data that carry semantic meanings.
Existing literature~\cite{bengio2013representation, lake2017building} demonstrates the potential of DRL in learning and understanding the world as humans do, 
where the understanding towards real-world observations can be reflected in disentangling the semantics in the form of disjoint factors. 
The disentanglement in the feature space encourages the learned representation to carry explainable semantics with independent factors, showing great potential to improve various machine learning tasks from the three aspects: i) Explainability: DRL learns semantically meaningful and separate representations which are aligned with latent generative factors. ii) Generalizability: DRL separates the representations that our tasks are interested in from the original entangled input and thus has better generalization ability. iii) Controllability: DRL achieves controllable generation by manipulating the learned disentangled representations in latent space.

\begin{figure}[htp]
    \centering
    \includegraphics[width=0.99\linewidth]{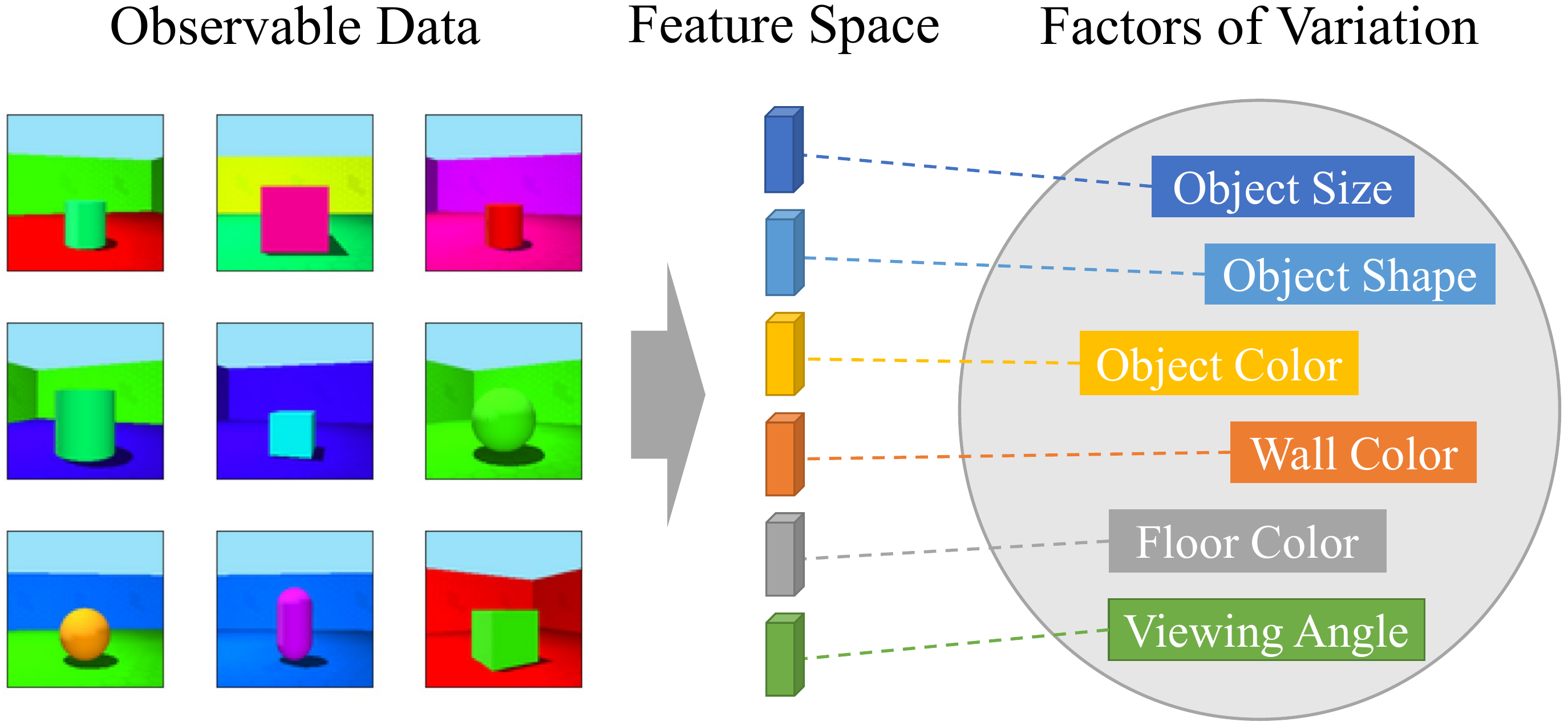}
    \caption{The scene of Shape3D~\cite{3dshapes18}, where the six rectangles in the gray circle represent the six factors of variation in the Shape3D respectively. DRL is expected to encode these distinct factors with independent latent variables in the latent feature space.}
    \label{fig:intro}
\end{figure}

Then a natural question arises, {\it What are disentangled representations supposed to learn?} 
The answer may lie in the concept of disentangled representation proposed by Bengio et al.~\cite{bengio2013representation}, which refers to \textit{factor of variations} in brief. 
As shown by the example illustrated in Figure~\ref{fig:intro}, Shape3D~\cite{3dshapes18} is a frequently used dataset in DRL with six distinct factors of variation, i.e., object size, object shape, object color, wall color, floor color and viewing angle. DRL aims at separating these factors and encoding them into independent and distinct latent variables in the representation space. In this case, the latent variables controlling object shape will 
change only with the variation of object shape and be constant over other factors. 
Analogously, it is the same for variables controlling other factors including size, color etc. 

Through both theoretical and empirical explorations, DRL benefits in the following three perspectives: i) Invariance: an element of the disentangled representations is invariant to the change of external semantics \cite{chen2019isolating, higgins2016beta, kim2018disentangling, Lee_2018_ECCV}, ii) Integrity: all the disentangled representations are aligned with real semantics respectively and are capable of generating the observed, undiscovered and even counterfactual samples \cite{chen2016infogan, larsen2016autoencoding, Yang_2021_CVPR, ghandeharioun2021dissect}, and  iii) Generalization: representations are intrinsic and robust instead of capturing confounded or biased semantics, thus being able to generalize for downstream tasks\cite{wu2020improving, suter2019robustly, lee2021learning}.

Following the motivation and requirement of DRL, there have been numerous works on DRL and its applications over various tasks.
Most typical methods for DRL are based on generative models~\cite{kingma2013auto, higgins2016beta, goodfellow2014generative, chen2016infogan}, which initially show great potential in learning explainable representations for visual images. 
In addition, approaches based on causal inference~\cite{suter2019robustly} and group theory~\cite{higgins2018towards} are widely adopted in DRL as well. 
The core concept of designing DRL architecture lies in encouraging the latent factors to learn disentangled representations while optimizing the inherent task objective, e.g., generation or discrimination objective.
Given the efficacy of DRL at capturing explainable, controllable and robust representations, it has been widely used in many fields such as computer vision~\cite{zhu2018visual, gonzalez2018image, Lee_2018_ECCV, Lee_2021_CVPR, Liu_2021_CVPR}, natural language processing~\cite{he2017unsupervised, bao2019generating, cheng2020improving}, recommender systems~\cite{ma2019learning, zhang2020content, wang2021multimodal, wang2020disentangled} and graph learning~\cite{ma2019disentangled, wang2020disentangled} etc., boosting the performances of various downstream tasks.

\textbf{Contributions.} In this paper, we comprehensively review DRL through summarizing the theories, methodologies, evaluations, applications and design schemes, to the best of our knowledge, for the first time. In particular, we present the definitions of DRL in Section~\ref{sec:definition} and comprehensively review DRL approaches in Section~\ref{sec:methods}. 
In Section~\ref{sec:metrics}, we discuss popular evaluation metrics for DRL implementation. 
We discuss the applications of DRL for various downstream tasks in Section~\ref{sec:applications}, followed by our insights in designing proper DRL models for different tasks in Section~\ref{sec:designs}. 
Last but not least, we summarize several open questions and future directions for DRL in Section~\ref{sec:future}.
Existing work most related to this paper is Liu et al.'s work~\cite{liu2022learning}, which only focuses on imaging domain and applications in medical imaging. In comparison, our work discusses DRL from a general perspective, taking full coverage of definitions, taxonomies, applications and design scheme.

\section{DRL Definitions}
\label{sec:definition}


\spara{Intuitive Definition}
Bengio et al.~\cite{bengio2013representation} propose an intuitive definition about disentangled representation:

\begin{definition}
\label{def:intuitive}
Disentangled representation should separate the distinct, independent and informative generative factors of variation in the data. Single latent variables are sensitive to changes in single underlying generative factors, while being relatively invariant to changes in other factors.
\end{definition}

The definition also indicates that latent variables are statistically independent. Following this intuitive definition, early DRL methods can be traced back to independent component analysis (ICA) and principal component analysis (PCA). Numerous Deep Neural Network (DNN) based methods also follow this definition~\cite{higgins2016beta,dupont2018learning,kim2019relevance,chen2016infogan,kim2018disentangling,chen2019isolating,burgess2018understanding,kumar2018variational,bouchacourt2018multi,bing2021disentanglement}. Most models and metrics hold the view that generative factors and latent variables are statistically independent. 

Definition~\ref{def:intuitive} is widely adopted in the literature, and is followed by the majority of DRL approaches discussed in Section~\ref{sec:methods}.

\begin{figure}[htbp]
\centering
   \begin{minipage}[t][40mm][t]{0.35\linewidth}
    \centering
    \includegraphics[width=\textwidth]{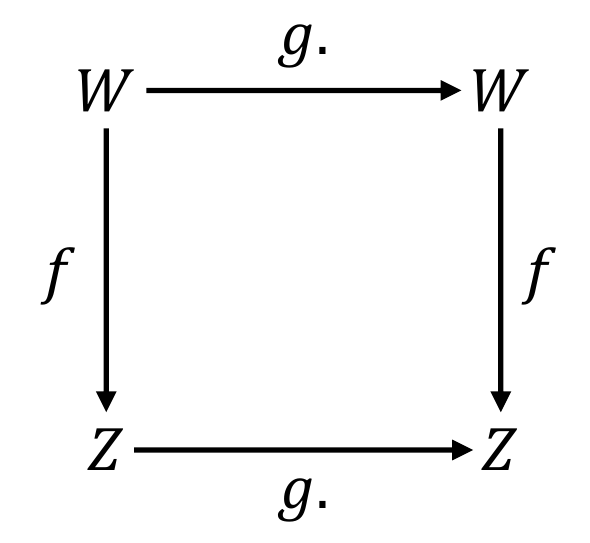}
    \caption{The illustration of condition (ii).}
    \label{fig:group_definition}
   \end{minipage} 
   \quad \qquad 
   \begin{minipage}[t]{0.45\linewidth}
    \centering
    \includegraphics[width=\textwidth]{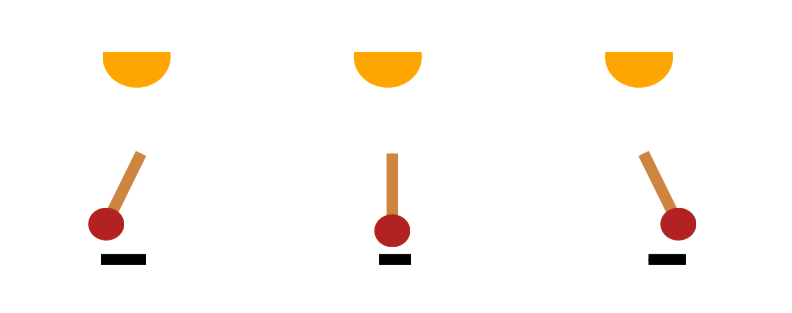}
    \caption{Swinging pendulum, light and shadow, figure from~\cite{Yang_2021_CVPR}.}
    \label{fig:pendulum}
   \end{minipage}
\end{figure}

\spara{Group Theory Definition}
For a more rigorous mathematical definition, Higgins et al.~\cite{higgins2018towards} propose to define DRL from the perspective of group theory, which is later adopted by a series of works~\cite{caselles2019symmetry,quessard2020learning,yang2021towards,wang2021self}. We briefly review the group theory-based definition as follows:

\begin{definition}
\label{def:group}
Consider a symmetry group $G$, world state space $W$ (i.e., ground truth factors which generate observations), data space $O$, and representation space $Z$. Assume $G$ can be decomposed as a direct product $G=G_{1} \times G_{2} \times \cdots \times G_{n}$. Representation $Z$ is disentangled with respect to $G$ if:

(i) There is an action of $G$ on $Z$: $G \times Z \rightarrow Z$.

(ii) There exists a mapping from $W$ to $Z$, i.e., $f: W\rightarrow Z$ which is equivariant between the action of $G$ on $W$ and $Z$. This condition can be formulated as follows:
\begin{align}
g \cdot f(w)=f(g \cdot w), \forall g \in G, \forall w \in W
\label{definition:1}
\end{align}
\noindent which can be illustrated as Figure.~\ref{fig:group_definition}.

(iii) The action of $G$ on $Z$ is disentangled with respect to the decomposition of $G$. In other words, there is a decomposition $Z=Z_{1} \times \ldots \times Z_{n}$ or $Z=Z_{1} \oplus \ldots \oplus Z_{n}$ such that each $Z_i$ is affected only by $G_i$ and invariant to $G_j, \forall j \neq i$.
\end{definition}

Definition~\ref{def:group} is mainly adopted by DRL approaches originating from the perspective of group theory in VAE (Group theory based VAEs in Section~\ref{method:VAE}).  

\spara{Discussions} All the two definitions hold the assumption that generative factors are naturally independent. 
However, Suter et al.~\cite{suter2019robustly} propose to define DRL from the perspective of the structural causal model (SCM)~\cite{pearl2009causality}, where they additionally introduce a set of confounders which causally influence the generative factors of observable data. Yang et al.~\cite{Yang_2021_CVPR} and Shen et al.~\cite{shen2020disentangled} further discard the independence assumption by considering that there might be an underlying causal structure which renders generative factors. For example, in Figure~\ref{fig:pendulum}, the position of the light source and the angle of the pendulum are both responsible for the position and length of the shadow. Consequently, instead of the independence assumption, they use SCM which characterizes the causal relationship of generative factors as prior. We refer to these works holding the assumption of causal factors as causal disentanglement methods, which will be discussed in detail in Section~\ref{ind vs. causal}.



\begin{figure*}[htp]
    \centering
    \includegraphics[width=1.0\linewidth]{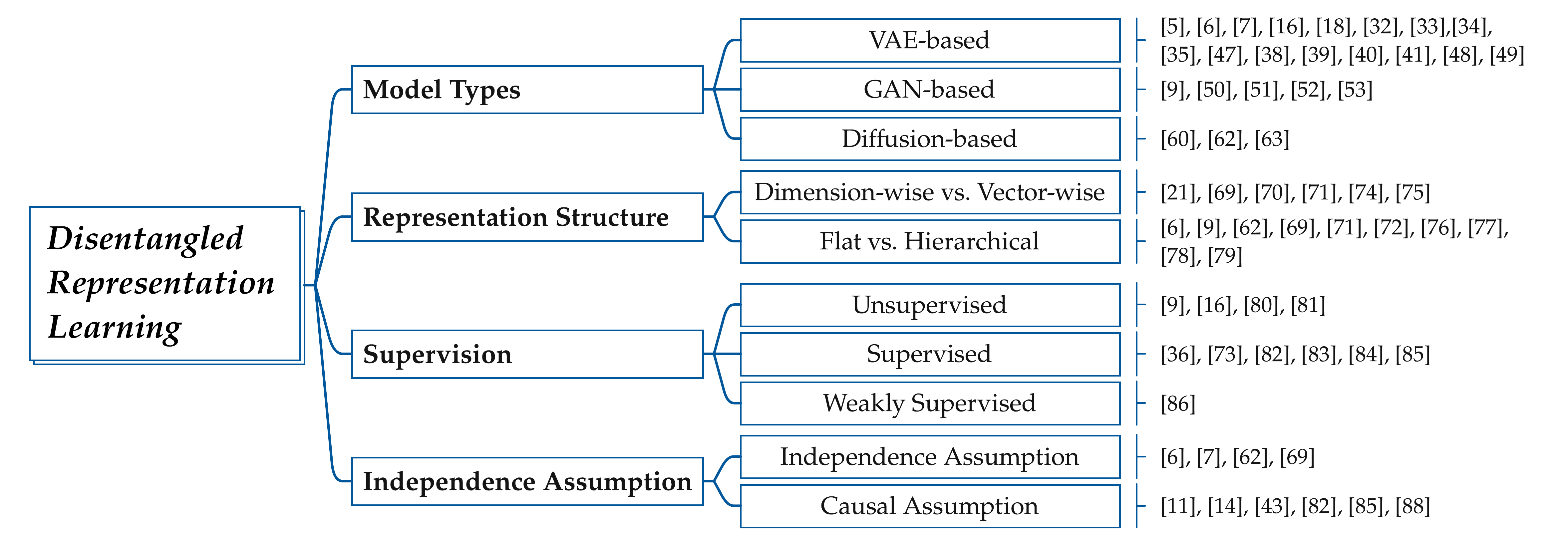}
    \caption{A categorization of DRL approaches.}
    \label{fig:taxonomy}
\end{figure*}

\section{DRL Taxonomy}
\label{sec:methods}


As shown in Figure~\ref{fig:taxonomy}, we categorize DRL approaches from \red{i) the perspective of base model type, ii) the perspective of representation structure, i.e., dimension-wise v.s. vector-wise and flat v.s. hierarchical, iii) the perspective of available supervision signal, i.e., unsupervised v.s. supervised v.s. weakly supervised, and iv) from the independence assumption of generative factors, i.e., independent v.s. causal.}
For each group, we will elaborate on specific models, show their advantages and disadvantages, as well as analyze their application scenarios.
Moreover, we also discuss \textit{Capsule Networks} and \textit{Object-centric Learning}, given that the two learning paradigms also employ the idea of disentanglement and thus can be regarded as particular instances of DRL.
We overview existing DRL methods and present inspirations on how DRL can be incorporated into specific tasks.

\subsection{Model Type}

\red{In this section, we classify DRL methods according to the base model architecture they use.
During the early investigation period of DRL, researchers commonly adopt VAEs or GANs as the backbone, 
and we refer to them as VAE-based or GAN-based DRL methods, which can be regarded as conventional methods. 
Recent studies have empowered DRL with more flexibilities, being capable of adapting to more advanced architectures such as Diffusion Models (DMs). 
We hope to emphasize the superiority and generalization of DRL by demonstrating its effectiveness in both conventional and advanced models. We will also provide some comparisons for these models.}

\subsubsection{VAE-based DRL Methods}
\label{method:VAE}
\noindent \textbf{Vanilla VAE-based Methods.}
Variational auto-encoder (VAE)~\cite{kingma2013auto} is a variant of the auto-encoder, which adopts the idea of variational inference. VAE is originally proposed as a deep generative probabilistic model for image generation. Later researchers find that VAE also has the potential ability to learn disentangled representation on simple datasets (e.g., FreyFaces~\cite{kingma2013auto}, MNIST~\cite{726791}). To obtain better disentanglement performance, researchers design various extra regularizers to combine with the original VAE loss function, resulting in the family of VAE-based Approaches. One of the most important characteristics of various VAE-based methods is the dimension-wise structure of the disentangled representations, i.e., different dimensions of the latent vector represent different factors. We will elaborate on the comparison of dimension-wise and the opposite vector-wise structure in Section~\ref{sec:dim vs. vec}.

\begin{figure}[htbp]
    \centering
    \includegraphics[width=0.9\linewidth]{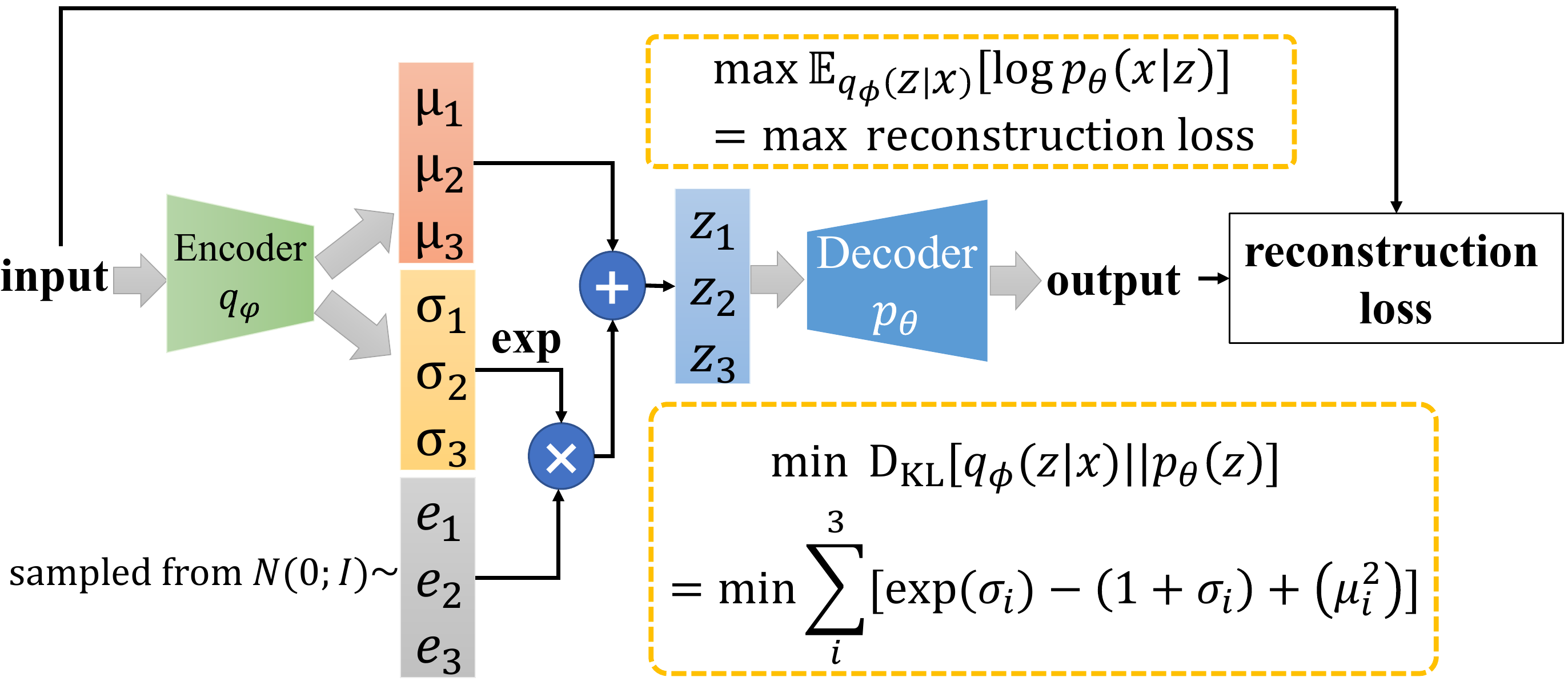}
    \caption{The general framework of variational auto-encoder (VAE).}
    \label{fig:vae}
\end{figure}

The general VAE model structure is shown in Figure~\ref{fig:vae}.
The fundamental idea of VAE is to model data distributions from the perspective of maximum likelihood using variational inference, i.e., to maximize $\log p_{\theta}(\mathbf{x})$. 
This objective can be written as Eq.\eqref{eq:vae1} in the following, 

{
\begin{align}
\small
\log p_{\theta}(\mathbf{x})=D_{K L}\big(q_{\phi}(\mathbf{z} | \mathbf{x}) \| p_{\theta}(\mathbf{z} | \mathbf{x})\big)+\mathcal{L}(\theta, \phi ; \mathbf{x},\mathbf{z}),
\label{eq:vae1}
\end{align}
\normalsize
}

\noindent where $q$ represents variational posterior distribution and $z$ represents the latent representation in hidden space. The key point of Eq.\eqref{eq:vae1} is leveraging variational posterior distribution $q_{\phi}(\mathbf{z}|\mathbf{x})$ to approximate true posterior distribution $p_{\phi}(\mathbf{z}|\mathbf{x})$, which is generally intractable in practice. The detailed derivation of Eq.\eqref{eq:vae1} can be found in the original paper~\cite{kingma2013auto}. The first term of Eq.\eqref{eq:vae1} is the KL divergence between variational posterior distribution $q_{\phi}(\mathbf{z}|\mathbf{x})$ and true posterior distribution $p_{\theta}(\mathbf{z}|\mathbf{x})$, and the second term is denoted as the (variational) evidence lower bound (ELBO) given that the KL divergence term is always non-negative. 
In practice, we usually maximize the ELBO to provide a tight lower bound for the original $\log(p_{\theta}(\mathbf{x}))$. 
The ELBO can also be rewritten as Eq.\eqref{eq:vae3} in the following,

{
\begin{align}
\small
\mathcal{L}(\theta,\phi;\mathbf{x},\mathbf{z})=&-D_{K L}\big(q_{\phi}(\mathbf{z} | \mathbf{x}) \| p_{\theta}(\mathbf{z})\big) \nonumber \\
&+\mathbb{E}_{q_{\phi}(\mathbf{z}| \mathbf{x})}\big[\log p_{\theta}(\mathbf{x} | \mathbf{z})\big],
\label{eq:vae3}
\end{align}
\normalsize
}

\noindent where 
the conditional logarithmic likelihood  $\mathbb{E}_{q_{\phi}(\mathbf{z}| \mathbf{x})}[\log p_{\theta}(\mathbf{x} | \mathbf{z})]$ is in charge of the reconstruction, and the KL divergence reflects the distance between the variational posterior distribution $q_{\phi}(\mathbf{z}|\mathbf{x})$ and the prior distribution $p_{\theta}(\mathbf{z})$. Generally, a standard Gaussian distribution $N(0,I)$ is chosen for $p_{\theta}(\mathbf{z})$ so that the KL term actually imposes independent constraints on the representations learned through neural network~\cite{chen2019isolating}, which may be the reason that VAE has the potential ability of disentanglement.

Although having the potential ability to disentangle, it has been observed that the vanilla VAE shows poor disentanglement capability on relatively complex datasets such as CelebA~\cite{liu2015faceattributes} and 3D Chairs~\cite{Aubry14} etc. 
To tackle this problem, a large amount of improvement has been proposed through adding implicit or explicit inductive bias to enhance disentanglement ability, resorting to various regularizers (e.g., ${\beta}$-VAE~\cite{higgins2016beta}, DIP-VAE~\cite{kumar2018variational}, and ${\beta}$-TCVAE~\cite{chen2019isolating} etc.). 
Specifically, to strengthen the independence constraint of the variational posterior distribution $q_{\phi}(\mathbf{z}|\mathbf{x})$, ${{\beta}}$-VAE~\cite{higgins2016beta} introduces a ${\beta}$ penalty coefficient before the KL term in ELBO, where the updated objective function is shown in Eq.\eqref{eq:beta1}. 

{\small
\begin{align}
\mathcal{L}(\theta, \phi, \mathbf{x}, \mathbf{z}, \beta)=
\mathbb{E}_{q_{\phi}(\mathbf{z} | \mathbf{x})}\big[\log p_{\theta}(\mathbf{x} | \mathbf{z})\big]-\beta D_{K L}\big(q_{\phi}(\mathbf{z} | \mathbf{x}) \| p_{\theta}(\mathbf{z})\big).
\label{eq:beta1}
\end{align}
}\normalsize

\noindent When ${\beta}$=1, ${\beta}$-VAE degenerates to the original VAE formulation. The experimental results of ${\beta}$-VAE~\cite{higgins2016beta} show that larger values of ${\beta}$ encourage learning more disentangled representations while harming the performance of reconstruction. 
Therefore, it is important to select an appropriate ${\beta}$ to control the trade-off between reconstruction accuracy and the quality of disentangling latent representations. To further investigate this trade-off phenomenon, Chen et al.~\cite{chen2019isolating} gives a more straightforward explanation from the perspective of ELBO decomposition. They prove that the penalty tends to increase dimension-wise independence of representation $\mathbf{z}$ but 
decrease the ability of $\mathbf{z}$ in preserving the information from input $\mathbf{x}$.

However, it is practically intractable to obtain the optimal $\beta$ that balances the trade-off between reconstruction and disentanglement. 
To handle this problem, Burgess et al.~\cite{burgess2018understanding} propose a simple modification, such that the quality of disentanglement can be improved as much as possible without losing too much information of the original data. 
They regard ${\beta}$-VAE objective as an optimization problem from the perspective of information bottleneck theory, whose 
objective function is shown in Eq.\eqref{eq:IB1} as follows,

{
\begin{align}
\small
\max [I(Z ; Y)-\beta I(X ; Z)],
\label{eq:IB1}
\end{align}
}\normalsize

\noindent where $X$ represents the original input to be compressed, $Y$ represents the objective task, $Z$ is the compressed representations for $X$, and 
$I( ; )$ stands for mutual information. Recall the ${\beta}$-VAE framework, we can regard the first term in Eq.\eqref{eq:beta1}, $\mathbb{E}_{q_{\phi}(\mathbf{z}| \mathbf{x})}[\log p_{\theta}(\mathbf{x} | \mathbf{z})]$ 
as $I(Z ; Y)$, and approximately treat the second term, $D_{K L}(q_{\phi}(\mathbf{z} | \mathbf{x}) \| p_{\theta}(\mathbf{z})\big)$ as $I(X ; Z)$.
To be specific, $q_{\phi}(\mathbf{z}|\mathbf{x})$ can be considered as the information bottleneck of the reconstruction task $\max \mathbb{E}_{q_{\phi}(\mathbf{z} | \mathbf{x})}[\log p_{\theta}(\mathbf{x} | \mathbf{z})]$. 
$D_{K L}\big(q_{\phi}(\mathbf{z} | \mathbf{x}) \| p_{\theta}(\mathbf{z})\big)$ can be seen as an upper bound over the amount of information that $q_{\phi}(\mathbf{z}|\mathbf{x})$ can extract and preserve for original data $\mathbf{x}$. The strategy is to gradually increase the information capacity of the latent channel, and the modified objective function is shown in Eq.\eqref{eq:understanding_bae} as follows,

{\small
\begin{align}
\mathcal{L}(\theta, \phi, C ; \mathbf{x}, \mathbf{z})=&\mathbb{E}_{q_{\phi}(\mathbf{z} \mid \mathbf{x})} \log p_{\theta}(\mathbf{x} | \mathbf{z})-  \nonumber \\
&\gamma\big|D_{K L}\big(q_{\phi}(\mathbf{z} | \mathbf{x}) \| p_{\theta}(\mathbf{z})\big)-C\big|,
\label{eq:understanding_bae}
\end{align}
}\normalsize

\noindent where $\gamma$ and $C$ are hyperparameters. During the training process, $C$ will gradually increase from $0$ to a value large enough to guarantee the expressiveness of latent representations, or in other words, to guarantee satisfactory reconstruction quality when achieving good disentanglement quality.

Furthermore, DIP-VAE~\cite{kumar2018variational} proposes an extra regularizer to improve the ability to disentangle, with objective function shown in Eq.\eqref{eq:dip-vae} as follows,

{\small
\begin{align}
\max_{\theta, \phi} \mathbb{E}_{\mathbf{x}}\Big[\mathbb{E}_{\mathbf{z} \sim q_{\phi}(\mathbf{z} | \mathbf{x})}\big[\log p_{\theta}(\mathbf{x} | \mathbf{z})\big]-&D_{K L}\big(q_{\phi}(\mathbf{z} | \mathbf{x}) \| p_{\theta}(\mathbf{z})\big)\Big] \nonumber \\
-&\lambda D\big(q_{\phi}(\mathbf{z}) \| p_{\theta}(\mathbf{z})\big),
\label{eq:dip-vae}
\end{align}
}\normalsize

\noindent where $D(\cdot \| \cdot)$ represents distance function between $q_{\phi}(\mathbf{z})$ and $p_{\theta}(\mathbf{z})$. 
The authors point out that $q_{\phi}(\mathbf{z})$ should equal to $\prod_{j} q_{j}\left(\mathbf{z}_{j}\right)$ to guarantee the disentanglement. Given the assumption that $p_{\theta}(\mathbf{z})$ follows the standard Gaussian distribution $N(0,I)$, the objective imposes independence constraint on the variational posterior cumulative distribution $q_{\phi}(\mathbf{z})$. In order to minimize the distance term, Kumar~et al. match the covariance of $q_{\phi}(\mathbf{z})$ and $p_{\theta}(\mathbf{z})$ by decorrelating the dimensions of $\mathbf{z} \sim q_{\phi}(\mathbf{z})$ given $p_{\theta}(\mathbf{z}) \sim N(0,I)$, i.e., they force Eq.\eqref{dip-vae:cov} to be close to the identity matrix,

{\small
\begin{align}
\operatorname{Cov}_{q_{\phi}(\mathbf{z})}[\mathbf{z}]=\mathbb{E}_{p(\mathbf{x})}\left[\boldsymbol{\Sigma}_{\phi}(\mathbf{x})\right]+\operatorname{Cov}_{p(\mathbf{x})}\left[\boldsymbol{\mu}_{\phi}(\mathbf{x})\right],
\label{dip-vae:cov}
\end{align}
}\normalsize

\noindent where $\boldsymbol{\mu}_{\phi}(\mathbf{x})$ and $\boldsymbol{\Sigma}_{\phi}(\mathbf{x})$ denote the prediction of VAE model for posterior $q_{\phi}(\mathbf{z}|\mathbf{x})$, i.e., $q_{\phi}(\mathbf{z}|\mathbf{x}) \sim N(\boldsymbol{\mu}_{\phi}(\mathbf{x}), \boldsymbol{\Sigma}_{\phi}(\mathbf{x}))$. 
Finally, they propose two variants, DIP-VAE-I and DIP-VAE-II, whose objective functions are shown in Eq.\eqref{dip-vae:2} and Eq.\eqref{dip-vae:3} respectively as follows, 

{\small
\begin{align}
\max_{\theta, \phi} \mathrm{ELBO}(\theta, \phi)&-\lambda_{o d} \sum_{i \neq j}\big[\operatorname{Cov}_{p(\mathbf{x})}[\boldsymbol{\mu}_{\phi}(\mathbf{x})]\big]_{i j}^{2}  \nonumber \\
&-\lambda_{d} \sum_{i}\Big(\big[\operatorname{Cov}_{p(\mathbf{x})}[\boldsymbol{\mu}_{\phi}(\mathbf{x})]\big]_{i i}-1\Big)^{2},
\label{dip-vae:2}
\end{align}
}

{\small
\begin{align}
\max _{\theta, \phi} \operatorname{ELBO}(\theta, \phi)&-\lambda_{o d} \sum_{i \neq j}\big[\operatorname{Cov}_{q_{\phi}(\mathbf{z})}[\mathbf{z}]\big]_{i j}^{2} \nonumber \\
&-\lambda_{d} \sum_{i}\Big(\big[\operatorname{Cov}_{q_{\phi}(\mathbf{z})}[\mathbf{z}]\big]_{i i}-1\Big)^{2},
\label{dip-vae:3}
\end{align}
}

\noindent where $\lambda_{d}$ and $\lambda_{od}$ are hyperparameters.
DIP-VAE-I regularizes $\operatorname{Cov}_{p(\mathbf{x})}\left[\boldsymbol{\mu}_{\phi}(\mathbf{x})\right]$, while DIP-VAE-II directly regularizes $\operatorname{Cov}_{q_{\phi}(\mathbf{z})}[\mathbf{z}]$.

Kim et al.~\cite{kim2018disentangling} propose FactorVAE which imposes independence constraint according to the definition of independence, as shown in Eq.\eqref{eq:factor},

{\small
\begin{align}
\frac{1}{N} \sum_{i=1}^{N}\Big[\mathbb{E}_{q_{\phi}(\mathbf{z} | \mathbf{x}^{(i)})}\big[\log p_{\theta}(\mathbf{x}^{(i)} | \mathbf{z})\big]&-D_{K L}(q_{\phi}(\mathbf{z} | \mathbf{x}^{(i)}) \| p_{\theta}\big(\mathbf{z})\big)\Big]  \nonumber \\
&-\gamma D_{K L}\big(q_{\phi}(\mathbf{z}) \| \bar{q}_{\phi}(\mathbf{z})\big),
\label{eq:factor}
\end{align}
}

\noindent where $\bar{q}_{\phi}(\mathbf{z})=\prod_{j} q_{\phi}\left(\mathbf{z}_{j}\right)$ and $\mathbf{x}^{(i)}$ represents $i$-th sample. 
${D_{K L}\big(q_{\phi}(\mathbf{z}) \| \prod_{j} q_{\phi}(\mathbf{z}_{j})\big)}$ is called \textit{Total Correlation} which evaluates the degree of dimension-wise independence in $\mathbf{z}$.

Chen et al.~\cite{chen2019isolating} propose to elaborately decompose $D_{K L}\big(q_{\phi}(\mathbf{z}|\mathbf{x})||p_{\theta}(\mathbf{z})\big)$ into three terms, as is shown in Eq.\eqref{eq:iso1}. i) The first term demonstrates the mutual information which can be rewritten as $I_q(\mathbf{z};\mathbf{x})$, ii) the second term denotes the total correlation and iii) the third term is the dimension-wise KL divergence. 

{\small
\begin{align}
D_{K L}(q_{\phi}\big(\mathbf{z}|\mathbf{x}) \| p_{\theta}(\mathbf{z})\big)=&\underbrace{D_{K L}\big(q_{\phi}(\mathbf{z},\mathbf{x}) \| q_{\phi}(\mathbf{z}) p_{\theta}(\mathbf{x})\big)}_{\text {(i) Mutual Information }} \nonumber
\\ 
&+\underbrace{D_{K L}\big(q_{\phi}(\mathbf{z}) \| \prod_{j} q_{\phi}(z_{j})\big)}_{\text {(ii) Total Correlation}} 
\nonumber
\\ 
&+\underbrace{\sum_{j} D_{K L}\big(q_{\phi}(z_{j}) \| p_{\theta}(z_{j})\big)}_{\text {(iii) Dimension-wise K L Divergence}}.
\label{eq:iso1}
\end{align}
}

\noindent From Eq.\eqref{eq:iso1}, we can straightforwardly obtain the explanation of the trade-off in ${\beta}$-VAE, i.e., higher ${\beta}$ tends to decrease $I_q(\mathbf{z};\mathbf{x})$ which is related to the reconstruction quality, while increasing the independence in $q_{\phi}(\mathbf{z})$ which is related to disentanglement. 
As such, instead of penalizing $D_{K L}\big(q_{\phi}(\mathbf{z}|\mathbf{x})||p_{\theta}(\mathbf{z})\big)$ as a whole with coefficient ${\beta}$, we can penalize these three terms with three different coefficients respectively, which is referred as ${\beta}$-TCVAE and is shown in Eq.\eqref{eq:iso2} as follows.

{\small
\begin{align}
\mathcal{L}=&\mathbb{E}_{q_{\phi}(\mathbf{z}|\mathbf{x}) p_{\theta}(\mathbf{x})}\big[\log p_{\theta}(\mathbf{x}|\mathbf{z})\big]-\alpha I_{q}(\mathbf{z};\mathbf{x})  \nonumber \\
&-\beta D_{K L}\big(q_{\phi}(\mathbf{z}) \| \prod_{j} q_{\phi}(z_{j})\big) 
- \gamma \sum_{j} D_{K L}\big(q_{\phi}(z_{j}) \| p_{\theta}(z_{j})\big).
\label{eq:iso2}
\end{align}
}

To further distinguish between meaningful and noisy factors of variation, Kim et al.~\cite{kim2019relevance} propose Relevance Factor VAE (RF-VAE) through introducing relevance indicator variables that are 
endowed with the ability to identify all meaningful factors of variation as well as the cardinality. The aforementioned VAE based methods are designed for continuous latent variables, failing to model the discrete variables. Dupont et al.~\cite{dupont2018learning} propose a ${\beta}$-VAE based framework, JointVAE, which is capable of disentangling both continuous and discrete representations in an unsupervised manner.  The formulas of the two methods are supplemented in Tabel~\ref{tab:summary_of_VAE}.

We conclude that all the above VAE based approaches are unsupervised, with the common characteristic of adding extra regularizer(s), 
e.g., $D_{K L}\big(q_{\phi}(\mathbf{z})||p(\mathbf{z})\big)$~\cite{kumar2018variational} and Total Correlation~\cite{kim2018disentangling}, 
in addition to ELBO such that the disentanglement ability can be guaranteed. 
The summary of these unsupervised VAE based approaches is illustrated in Table~\ref{tab:summary_of_VAE}. Besides, there are also several works that incorporate supervised signals into VAE-based models, which we will discuss in the Section~\ref{unsupervised vs. supervised}.



\begin{table*}[h]
\small
	\caption{The summary of VAE based approaches.}
	\label{tab:summary_of_VAE}
	\centering
	\begin{tabular}{|m{3cm}<{\centering}|m{6cm}<{\centering}|m{8cm}|}\hline
			Method& Regularizer& Description \\\hline
			${\beta}$-VAE 
			&$-\beta D_{K L}\big(q_{\phi}(\mathbf{z} | \mathbf{x}) \| p(\mathbf{z})\big)$
			&${\beta}$ controls the trade-off between
            reconstruction fidelity and the quality of disentanglement in latent representations. 
			\\ \hline
			Understanding disentangling in ${\beta}$-vae
			&$-\gamma\left|D_{K L}\big(q_{\phi}(\mathbf{z} | \mathbf{x}) \| p(z)\big)-C\right|$
			& The quality of disentanglement can be improved as much as possible without losing too much information from original data by linearly increasing $C$ during training.
			\\ \hline
			DIP-VAE
		    & $-\lambda D_{K L}\big(q_{\phi}(\mathbf{z}) \| p(\mathbf{z})\big)$
		    & Enhance disentanglement by minimizing the distance between $q_{\phi}(\mathbf{z})$ and $p(\mathbf{z})$. In practice, we can match the moments between $q_{\phi}(\mathbf{z})$ and $p(\mathbf{z})$.
		    \\ \hline
		    FactorVAE
		    &${-{\gamma}D_{K L}\big(q_{\phi}(\mathbf{z}) \| \prod_{j} q_{\phi}(z_{j})\big)}$
		    &Directly impose independence constraint on $q_{\phi}(\mathbf{z})$ in the form of total correlation.
		    \\ \hline
		     $\beta$-TCVAE
		    &$-\alpha I_{q}(\mathbf{z};\mathbf{x})-\beta D_{K L}\big(q(\mathbf{z}) \| \prod_{j} q(z_{j})\big) \nonumber 
            -\gamma \sum_{j} D_{K L}\big(q(z_{j}) \| p(z_{j})\big)$
            &Decompose $D_{K L}\big(q(\mathbf{z}|\mathbf{x})||p(\mathbf{z})\big)$ into three terms: i) mutual information, ii) total correlation, iii) dimension-wise KL divergence and then penalize them respectively.
            \\ \hline
            JointVAE
            &$-\gamma\left|D_{K L}\big(q_{\phi}(\mathbf{z} | \mathbf{x}) \| p(\mathbf{z})\big)-C_{z}\right|
            -\gamma\left|D_{K L}\big(q_{\phi}(\mathbf{c} | \mathbf{x}) \| p(\mathbf{c})\big)-C_{c}\right|$
            &Separate latent variables into continuous $\mathbf{z}$ and discrete $\mathbf{c}$, then modify the objective function of ${\beta}$-VAE to capture discrete generative factors. 
            \\ \hline
            RF-VAE
            &$-\sum_{j=1}^{d} \lambda\left(r_{j}\right) D_{K L}\big(q(z_{j} | \mathbf{x}) \| p(z_{j})\big)-\gamma D_{K L}\big(q(\mathbf{r} \circ \mathbf{z}) \| \prod_{j=1}^{d} q(r_j \circ z_{j})\big) -\eta\|\mathbf{r}\|_{1}
            $
            &Introduce relevance indicator variables $\mathbf{r}$ by only focusing on relevant part when computing the total correlation,
            penalize $D_{K L}\big(q(z_j|\mathbf{x})||p(z_j)\big)$ less for relevant dimensions and more for nuisance (noisy) dimensions.
            \\ \hline
	\end{tabular}
\end{table*}

VAE-based methods can also be modified to process sequential data, e.g., video and audio. Li et al.~\cite{li2018disentangled} separate latent representations of video frames into time-invariant and time-varying parts. They use $\mathbf{f}$ to model the global time-invariant aspects of the video frames, and use $\mathbf{z_i}$ to represent the time-varying feature of the $i$-th frame. The training procedure conforms to the VAE algorithm~\cite{kingma2013auto} with the objective of maximizing ELBO in Eq.\eqref{eq:disentangle_vae_4} as follows,






{\small
\begin{align}
\mathcal{L}(\theta, \phi, \mathbf{x}, \mathbf{z}, \mathbf{f})=
&\mathbb{E}_{q_{\phi}(\mathbf{z_{1: T}},\mathbf{f} | \mathbf{x_{1: T}})}\big[\log p_{\theta}(\mathbf{x_{1: T}} | \mathbf{z_{1: T}},\mathbf{f})\big] \nonumber \\ 
&-D_{K L}\big(q_{\phi}(\mathbf{z_{1: T}},\mathbf{f} | \mathbf{x_{1: T}}) \| p_{\theta}(\mathbf{z_{1: T}},\mathbf{f})\big)
\label{eq:disentangle_vae_4}
\end{align}
}

\noindent \textbf{VAE-based Methods with Group Theory.} Besides the intuitive definition from Definition~\ref{def:intuitive}, Higgins et al.~\cite{higgins2018towards} propose a mathematically rigorous group theory definition of DRL in Definition~\ref{def:group}, which is followed by a series of works~\cite{caselles2019symmetry,quessard2020learning,yang2021towards,wang2021self} on group-based DRL.

Quessard et al.~\cite{quessard2020learning} propose a method for learning disentangled representations of dynamical environments (which returns observations) 
from the trajectories of transformations (which act on the environment). They consider the data space $O$ and latent representation space $V$, where a dataset of trajectories $(o_0,g_0,o_1,g_1,...)$ with $o_i$ denoting the observation of data and $g_i \in G$ denoting the transformation that transforms $o_i$ to $o_{i+1}$. 
They map $G$ to a group of matrices belonging to the special orthogonal group $SO(n)$, i.e., mapping $g_i$ to an element of general linear group $GL(V)$, which is shown in Eq.\eqref{eq:group1}). $R_{i,j}$ denotes the rotation in the $(i,j)$ plane. For instance, in the case of 3-dimensional space:

{\small
\begin{align}
R_{i,i}\left(\theta_{i,j}\right)=\left(\begin{array}{ccc}
\cos \theta_{i,j} & 0 & \sin \theta_{i,j} \\
0 & 1 & 0 \\
-\sin \theta_{i,j} & 0 & \cos \theta_{i,j}
\end{array}\right).
\end{align}
}

\noindent In the training procedure, they first randomly select an observation $o_i$ in the trajectories, then generate a series of reconstructions $\{\hat{o}_{k}\}_{k=i+1, \ldots, i+m}$ through Eq.\eqref{eq:group2}, where $f_\phi$ is the encoder mapping the observations to the n-dimensional latent space $V$ and $d_{\psi}$ is the decoder. The first objective is to minimize the reconstruction loss $\mathcal{L}_{\mathrm{rec}}(\phi, \psi, \theta)$ between the true observations $\{{o}_{k}\}_{k=i+1, \ldots, i+m}$ generated by the transformations in the environment and the reconstructed observations $\{\hat{o}_{k}\}_{k=i+1, \ldots, i+m}$ generated by the transformations in the latent space. Furthermore, to enforce disentanglement, they propose another loss function $\mathcal{L}_{ent}(\theta)$ which penalizes the number of rotations that a transformation $g_a(\theta^{a}_{i,j})$ involves, which is shown in Eq.\eqref{eq:group3}. Lower $\mathcal{L}_{ent}$ indicates that $g_a$ involves fewer rotations and thus $g_a$ acts on fewer dimensions, which means better disentanglement.

{\small
\begin{align}
g\left(\theta_{1,2}, \theta_{1,3} ..., \theta_{1, n}, \theta_{2, 3} ..., \theta_{2, n},...... \theta_{n-1, n}\right)=\prod_{i=1}^{n-1} \prod_{j=i+1}^{n} R_{i, j}\left(\theta_{i, j}\right) ,
\label{eq:group1}
\end{align}
}

{\small
\begin{align}
\hat{o}_{i+m}(\phi, \psi, \theta)=d_{\psi}\big(g_{i+m}(\theta) \cdot g_{i+m-1}(\theta) \ldots . g_{i+1}(\theta) \cdot f_{\phi}(o_{i})\big) ,
\label{eq:group2}
\end{align}
}

{\small
\begin{align}
\mathcal{L}_{\mathrm{ent}}(\theta)=\sum_{a} \sum_{(i, j) \neq(\alpha, \beta)}\left|\theta_{i, j}^{a}\right|^{2} \quad \text { with } \quad \theta_{\alpha, \beta}^{a}=\max _{i, j}\left(\left|\theta_{i, j}^{a}\right|\right) .
\label{eq:group3}
\end{align}
}

Different from environment-based methods~\cite{quessard2020learning,caselles2019symmetry} which leverage environment to provide world states, Yang et al.~\cite{yang2021towards} propose a theoretical framework to make Definition~\ref{def:group} feasible in the setting of unsupervised DRL without relying on the environment. 
They propose three sufficient conditions in the framework, namely model constraint, data constraint and group structure constraint, together with a specific implementation of the framework based on the existing VAE-based models through integrating additional loss. 
The authors assume that $G$ is a direct product of $m$ rings of integers modulo $n$, i.e., $G=(\mathbb{Z} / n \mathbb{Z})^{m}=\mathbb{Z} / n \mathbb{Z} \times \mathbb{Z} / n \mathbb{Z} \times \cdots \times \mathbb{Z} / n \mathbb{Z}$, where $n$ denotes the number of possible values for a factor and $m$ denotes the number of all factors. They assume $Z$ has the same elements as $G$ and further assume the group action of $G$ on $Z$ is element-wise addition, i.e., $g \cdot z=\overline{g+z}, \forall z \in Z, g \in G$. In order not to involve the group action on world state space $W$ for the unsupervised setting, they construct the permutation group $\Phi$, then use the group action of $\Phi$ on the data space $O$ to replace the group action of $G$ on $W$, which can be formulated in Eq.\eqref{eq:group4} as follows,

{\small
\begin{align}
f(g \cdot w)=h\big(\varphi_{g} \cdot b(w)\big)=h\left(\varphi_{g} \cdot o\right), \forall w \in W, g \in G ,
\label{eq:group4}
\end{align}
}

\noindent where $h$ represents the mapping from $O$ to $Z$ and $b$ represents the mapping from $W$ to $O$.

Here the $\Phi$ satisfying Eq.\eqref{eq:group4} exists if and only if: 
(i) $\Phi$ is isomorphic to $G$; 
(ii) For each generator of dimension $i$ of $G$, i.e., $g_i$, there exists a generator of $\Phi$, i.e., $\varphi_{i}$, such that $\varphi_{i} \cdot b(w)=b\left(g_{i} \cdot w\right), \forall w \in W$;
(iii) $\varphi_{g} \cdot b(w)=h^{-1}\big(g \cdot f(w)\big), \forall w \in W, \varphi_{g} \in \Phi$, where $\varphi_{g}$ is the corresponding element of $g$ under the isomorphism.
Condition (i) and (ii) are respectively referred as group structure constraint and data constraint. 
Condition (iii) is a model constraint which further guarantees that group $\Phi$ can be achieved by encoder, decoder and the group action of $G$ on $Z$. When these three conditions are satisfied, it can be derived that $Z$ is disentangled with respect to $G$. 
However, given that condition (ii) directly involves the world states, a learning method named GROUPIFIED VAE utilizing a necessary condition to 
substitute (ii) is proposed to satisfy the unsupervised setting. 
Thus under the architecture of VAE, the model constraint in condition (iii) can be formulated by Eq.\eqref{eq:group5} as follows,

{\small
\begin{align}
\varphi_{g} \cdot o=h^{-1}\big(g \cdot h(o)\big) \triangleq d\big(g \cdot h(o)\big), \forall o \in O, g \in G,
\label{eq:group5}
\end{align}
}

\noindent where $h$ is the encoder and $d$ is the decoder. 
Moreover, the data constraint can be satisfied to some extent by VAE based models for the unsupervised setting because of the intuition that VAE based models can generate the data from statistical independent latent variables which are similar to generators of $\Phi$.  To satisfy the group structure constraint, GROUPIFIED VAE proposes \textit{Abel Loss} and \textit{Order Loss} to guarantee that $\phi$ is isomorphic to $G$, which are formulated in Eq.\eqref{eq:group6} and Eq.\eqref{eq:group7} respectively as follows,

{\small
\begin{align}
\mathcal{L}_{a}=\sum_{o \in O} \sum_{ (i, j)}\big\|\varphi_{i} \cdot\left(\varphi_{j} \cdot o\right)-\varphi_{j} \cdot\left(\varphi_{i} \cdot o\right)\big\|,
\label{eq:group6}
\end{align}
}
{\small
\begin{align}
\mathcal{L}_{o}=\sum_{o \in O} \sum_{1 \leq i \leq m}\big(\left\|\varphi_{i} \cdot\left(\varphi_{i}^{n-1} \cdot o\right)-o\right\|+\left\|\varphi_{i}^{n-1} \cdot\left(\varphi_{i} \cdot o\right)-o\right\|\big),
\label{eq:group7}
\end{align}
}
\noindent where $\varphi$ is the generator of $\Phi$.

Beyond learning the homomorphism from a group to group action, Wang et al.~\cite{wang2021self} propose Iterative Partition-based Invariant Risk Minimization (IP-IRM), an iterative algorithm based on the self-supervised learning fashion, to specifically learn a mapping between observation space $\mathcal{I}$ and feature space $\mathcal{X}$, i.e., a disentangled feature extractor $\phi$ such that $\mathbf{x}=\phi(I)$ under the group-theoretical disentanglement conditions. They first argue that most existing self-supervised learning approaches only disentangle the augmentation related features, thus failing to modularize the global semantics. In contrast, IP-IRM is able to ground the abstract semantics and the group actions successfully.
Specifically, IP-IRM partitions the training data into disjoint subsets with a partition matrix $\mathbf{P}$, and defines a pretext task with contrastive loss $\mathcal{L}(\phi,\theta = 1, k, \mathbf{P})$ on the samples in the $k$-th subset, where $\theta$ is a constant parameter. At each iteration, it finds a new partition $\mathbf{P}^*$ through maximizing the variance across the group orbits by Eq.\eqref{eq:group8}, which reveals an entangled group element $g_i$.

{\small
\begin{align}
\mathbf{P}^* = \mathop{\arg\max}_\mathbf{P} & \sum_k \big[\mathcal{L}(\phi,\theta = 1, k, \mathbf{P}) \nonumber \\
 & + \lambda_2 \left\| \nabla_{\theta=1} \mathcal{L}(\phi,\theta = 1, k, \mathbf{P}) \right\|^2 \big].
\label{eq:group8}
\end{align}
}

Then the Invariant Risk Minimization (IRM)~\cite{arjovsky2019invariant} approach is adopted to update $\phi$ by Eq.\eqref{eq:group9}, which disentangles the representation w.r.t $g_i$. It sets $\mathcal{P}=\{\mathbf{P}\}$ at beginning and update $\mathcal{P} \leftarrow \mathcal{P} \cup \mathbf{P}^*$ each time.

{\small
\begin{align}
\min_\phi \sum_{\mathbf{P}\in\mathcal{P}} \sum_k \big[\mathcal{L}(\phi,\theta = 1, k, \mathbf{P}) + \lambda_1 \left\| \nabla_{\theta=1} \mathcal{L}(\phi,\theta = 1, k, \mathbf{P}) \right\|^2 \big]
\label{eq:group9}
\end{align}
}

It is theoretically proved that iterating the above two steps eventually converges to a fully disentangled representation w.r.t. $\prod_{i=1}^m g_i$. IP-IRM is devised to delay the group action learning to downstream tasks on demand so that it learns a disentangled representation with an inference process, which provides wide feasibility and availability on large-scale tasks.

Moreover, Zhu et al.~\cite{zhu2021commutative} propose an unsupervised DRL framework, named Commutative Lie Group VAE. They introduce a matrix Lie group $G$ and corresponding Lie algebra $\mathfrak{g}$ which satisfies Eq.\eqref{eq:group10},

{\small
\begin{align}
& g(t)=\exp (B(t)), g \in G, B \in \mathfrak{g} , \nonumber \\
& B(t)=t_{1} B_{1}+t_{2} B_{2}+\ldots+t_{m} B_{m}, \forall t_{i} \in \mathbb{R},
\label{eq:group10}
\end{align}
}

\noindent where exp($\cdot$) denotes the matrix exponential map and $\left\{B_{i}\right\}_{i=1}^{m}$ is a basis of the Lie algebra. 
In this case, every sample have a group representation $\mathbf{z}$ and can also be identified by coordinate $t$ in the Lie algebra. 
The objective function is written in Eq.\eqref{eq:group11} as follows,

{\small
\begin{align}
\log p(\mathbf{x}) \geq &\mathcal{L}_{\text {bottleneck }}(\mathbf{x}, \mathbf{z}, \mathbf{t}) \nonumber \\
=& \mathbb{E}_{q(\mathbf{z} | \mathbf{x}) q(\mathbf{t} | \mathbf{z})} \log p(\mathbf{x} | \mathbf{z}) p(\mathbf{z} | \mathbf{t}) \nonumber \\
&-\mathbb{E}_{q(\mathbf{z} | \mathbf{x})} D_{K L}\big(q(\mathbf{t} | \mathbf{z})|| p(\mathbf{t})\big)-\mathbb{E}_{q(\mathbf{z} | \mathbf{x})} \log q(\mathbf{z} | \mathbf{x}),
\label{eq:group11}
\end{align}
}

\noindent where $q(\mathbf{z}|\mathbf{x})$ is implemented as a deterministic encoder, while $q(\mathbf{t}|\mathbf{z})$ is implemented as a stochastic encoder. $p(\mathbf{z}|\mathbf{t})$ is implemented through $\mathbf{z}=g(\mathbf{t})$, and $p(\mathbf{x}|\mathbf{z})$ is implemented as an image decoder. 
The first and second term can be regarded as reconstruction loss on data space and representation space respectively. The third term is the conditional entropy, which is constant. Moreover, a one-parameter decomposition constraint and a Hessian penalty constraint on $\left\{B_{i}\right\}_{i=1}^{m}$ are proposed to encourage disentanglement as well.

\subsubsection{GAN-based DRL Methods}
\label{sec:gan-based method}
GAN (Generative Adversarial Net)~\cite{goodfellow2014generative}, as another important generative model proposed by Goodfellow et al., has drawn a lot of attention from researchers.
Instead of adopting conventional Bayesian statistical methods, GAN directly sample latent representations $\mathbf{z}$ from a prior distribution $p(\mathbf{z})$. 
Specifically, GAN has a generative network (generator) G and a discriminative network (discriminator) $D$ where the generator G simulates a complex unknown generative system which transforms latent representation $\mathbf{z}$ to a generated image, while the discriminator $D$ receives an image (real or generated by $G$) as input and then outputs the probability of the input image being real. 
In the training process, the goal of generator $G$ is to generate images which can deceive discriminator $D$ into believing the generated images are real. 
Meanwhile, the goal of discriminator $D$ is to distinguish the images generated by generator $G$ from the real ones. 
Thus, generator $G$ and discriminator $D$ constitute a dynamic adversarial \textit{minimax} game. Ideally, generator $G$ can finally generate an image that looks like a real one so that discriminator $D$ fails to determine whether the image generated by generator $G$ is real or not. 
The objective function is shown as Eq.\eqref{eq:gan1}, 

{\small
\begin{align}
\min_{G} \max_{D} V(D, G)&=\mathbb{E}_{\mathbf{x} \sim P_{data}}\big[\log D(\mathbf{x})\big] \nonumber \\
&+\mathbb{E}_{\mathbf{z} \sim p(\mathbf{z})}\Big[\log \Big(1-D\big(G(\mathbf{z})\big)\Big)\Big],
\label{eq:gan1}
\end{align}
}

\noindent where $P_{data}$ represents the real dataset and $p(\mathbf{z})$ represents the prior distribution of the latent representation $\mathbf{z}$.

Similar to VAE-based methods, researchers have explored a mass of GAN-based methods to achieve DRL.

InfoGAN~\cite{chen2016infogan} is one of the earliest works using the GAN paradigm to conduct dimension-wise DRL. The generator takes two latent variables as input, where one is the incompressible noise $\mathbf{z}$, and the other is the target latent variable $\mathbf{c}$ which captures the latent generative factors. To encourage the disentanglement in $\mathbf{c}$, InfoGAN designs an extra variational regularization of mutual information, i.e., $I(\mathbf{c} ; G(\mathbf{z}, \mathbf{c}))$ controlled by hyperparameter $\lambda$, such that the adversarial loss of InfoGAN is written in Eq.~\eqref{eq:infogan1} as follows,

{\small
\begin{align}
\min_{G} \max_{D} V_{I}(D, G)=V^{'}(D, G)-\lambda I\big(\mathbf{c} ; G(\mathbf{z}, \mathbf{c})\big),
\label{eq:infogan1}
\end{align}
}

\noindent where $V^{'}(D, G)$ is defined in Eq.\eqref{eq:infogan3}, taking $\mathbf{c}$ into account.

{\small
\begin{align}
 V^{\prime}(D, G)&=\mathbb{E}_{\mathbf{x} \sim P_{data }}\big[\log D(\mathbf{x})\big] \nonumber \\
&+\mathbb{E}_{\mathbf{z} \sim p(\mathbf{z})}\Big[\log \Big(1-D\big(G(\mathbf{z},\mathbf{c})\big)\Big)\Big].
\label{eq:infogan3}
\end{align}
}

\begin{figure}[htp]
    \includegraphics[width=0.99\linewidth]{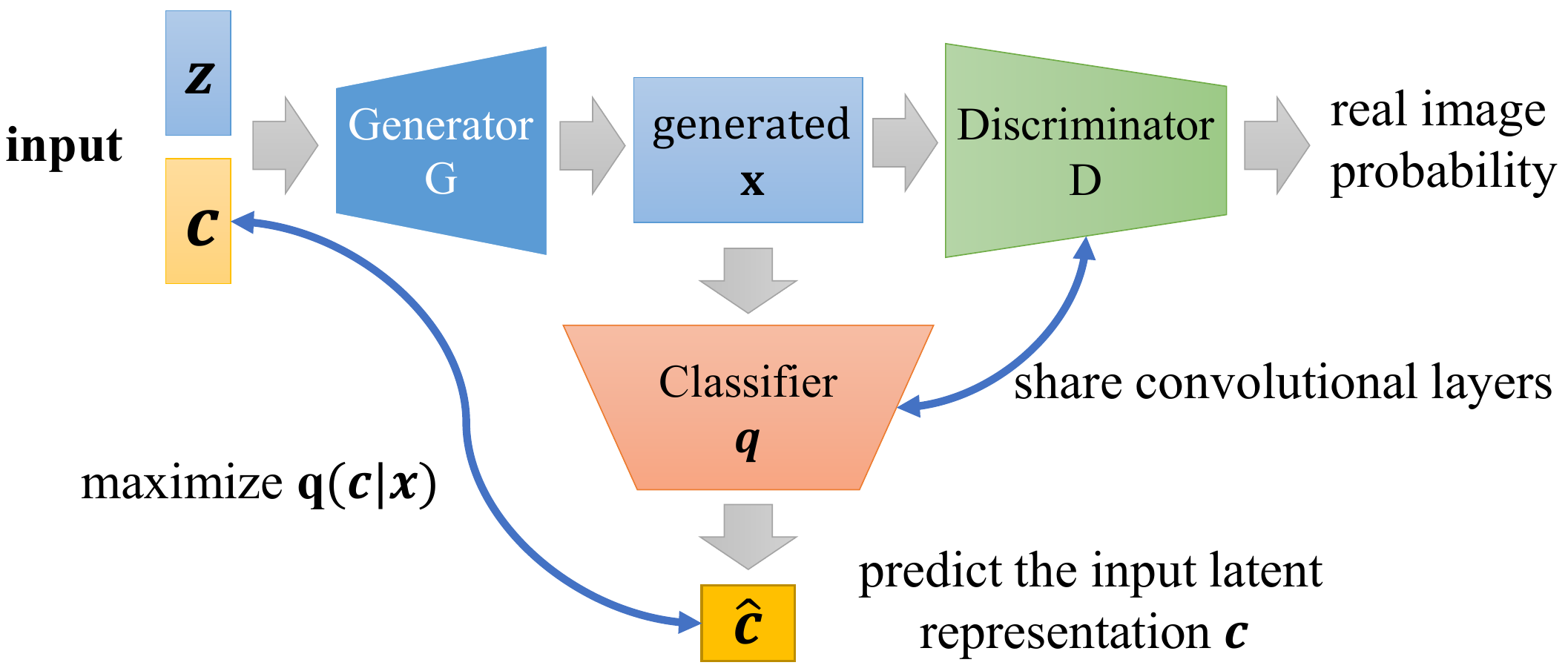}
    \caption{The overall framework of InfoGAN.}
    \label{fig:infogan}
\end{figure}

However, it is intractable to directly optimize $I\big(\mathbf{c} ; G(\mathbf{z}, \mathbf{c})\big)$ because of the inaccessibility of posterior $p(\mathbf{c}|\mathbf{x})$. Therefore, 
InfoGAN derives a lower bound of $I\big(\mathbf{c} ; G(\mathbf{z}, \mathbf{c})\big)$ with variational inference in Eq.\eqref{infogan:lower_bound}, 

{\small
\begin{align} 
I\big(\mathbf{c} ; G(\mathbf{z}, \mathbf{c})\big) & = H(\mathbf{c})-H\big(\mathbf{c} \mid G(\mathbf{z}, \mathbf{c})\big) \nonumber \\
& = \mathbb{E}_{\mathbf{x} \sim G(\mathbf{z}, \mathbf{c})}\Big[\mathbb{E}_{\mathbf{c}^{\prime} \sim p(\mathbf{c} \mid \mathbf{x})}\left[\log p\left(\mathbf{c}^{\prime} \mid \mathbf{x}\right)\right]\Big]+H(\mathbf{c})  \nonumber \\
& = \mathbb{E}_{\mathbf{x} \sim G(\mathbf{z}, \mathbf{c})}\Big[\underbrace{D_{K L}\big(p(\cdot \mid \mathbf{x}) \| q(\cdot \mid \mathbf{x})\big)}_{\geq 0} \nonumber \\
& \quad \quad +\mathbb{E}_{\mathbf{c}^{\prime} \sim p(\mathbf{c} \mid \mathbf{x})}\left[\log q\left(\mathbf{c}^{\prime} \mid \mathbf{x}\right)\right]\Big]+H(\mathbf{c}) \nonumber \\
& \geq \mathbb{E}_{\mathbf{x} \sim G(\mathbf{z}, \mathbf{c})}\Big[\mathbb{E}_{\mathbf{c}^{\prime} \sim p(\mathbf{c} \mid \mathbf{x})}\left[\log q\left(\mathbf{c}^{\prime} \mid \mathbf{x}\right)\right]\Big]+H(\mathbf{c}) \nonumber \\
& = \mathbb{E}_{\mathbf{c} \sim p(\mathbf{c}), \mathbf{x} \sim G(\mathbf{z}, \mathbf{c})}[\log q(\mathbf{c} \mid \mathbf{x})]+H(\mathbf{c}),
\label{infogan:lower_bound}
\end{align}
}
\noindent where $H(.)$ denotes the entropy of the random variable and $q(\mathbf{c}|\mathbf{x})$ is the auxiliary posterior distribution approximating the true posterior $p(\mathbf{c}|\mathbf{x})$. Actually, $q$ is implemented as a neural network. 
The overall framework of InfoGAN is shown in Figure.~\ref{fig:infogan}.

Nevertheless, the performance of InfoGAN for disentanglement is constantly reported to be lower than VAE-based models. To enhance disentanglement, Jeon et al.~\cite{jeon2021ib} propose IB-GAN which compresses the representation by adding a constraint on the maximization of mutual information between latent representation $\mathbf{z}$ and $G(\mathbf{z})$, which is actually a kind of application for information bottleneck. The hypothesis behind IB-GAN is that the compressed representations usually tend to be more disentangled. 

Lin et al.~\cite{lin2019infogan} propose InfoGAN-CR, which is a self-supervised variant of InfoGAN with contrastive regularizer. 
They generate multiple images by keeping one dimension of the latent representation, i.e., $c_i$, fixed and randomly sampling others, i.e., $c_j$ where $j\neq i$.  
Then a classifier which takes these images as input will be trained to determine which dimension is fixed. 
The contrastive regularizer encourages distinctness across different dimensions in the latent representation, thus being capable of promoting disentanglement.

Zhu et al.~\cite{zhu2021and} propose PS-SC GAN based on InfoGAN which employs a Spatial Constriction (SC) design to obtain the focused areas of each latent dimension and utilizes Perceptual Simplicity (PS) design to encourage the factors of variation captured by latent representations to be simpler and purer. The Spatial Constriction design is implemented as a spatial mask with constricted modification. Moreover, PS-SC GAN imposes a perturbation $\epsilon$ on a certain latent dimension $c_i $ (i.e., $c_{i}^{\prime}=c_{i}+\epsilon$) and then computes the reconstruction loss between $\mathbf{c}$ and $\hat{\mathbf{c}}$ with $\hat{\mathbf{c}}=q\big(G(\mathbf{c},\mathbf{z})\big)$, as well as the reconstruction loss between $\mathbf{c}^{\prime}$ and $\hat{\mathbf{c}}^{\prime}$ with
$\hat{\mathbf{c}}^{\prime}=q\big(G(\mathbf{c}^{\prime},\mathbf{z})\big)$, where $q$ is a classifier same in InfoGAN. The principle of Perceptual Simplicity is to punish more on the reconstruction errors for the perturbed dimensions and give more tolerance for the misalignment of the remaining dimensions.

Wei et al.~\cite{wei2021orthogonal} propose an orthogonal Jacobian regularization (OroJaR) to enforce disentanglement for generative models. They employ the Jacobian matrix of the output with respect to the input (i.e., latent variables for representation) to measure the output changes caused by the variations in the input.
Assuming that the output changes caused by different dimensions of latent representations are independent with each other, 
then the Jacobian vectors are expected to be orthogonal with each other, i.e., minimizing Eq.~\eqref{eq:jacobian_reg}, 

{\small
\begin{align}
\mathcal{L}_{\text{Jacob}}(G)=\sum_{d=1}^{D} \sum_{i=1}^{m} \sum_{j \neq i}\left|\left[\frac{\partial G_{d}}{\partial z_{i}}\right]^{T} \frac{\partial G_{d}}{\partial z_{j}}\right|^{2},
\label{eq:jacobian_reg}
\end{align}
}

\noindent where $G_d$ denotes the $d$-th layer of the generative models and $z_i$ denotes $i$-th dimension in the latent representation.



\subsubsection{Diffusion-based DRL Methods}
\label{sec: diffusion methods}

\red{On the one hand, diffusion model (DM) is an advanced class of generative model which has drawn a lot of attention in recent years. Diffusion model shows remarkable performance in various tasks, including text-to-image generation~\cite{rombach2022high}, image editing~\cite{kawar2023imagic}, and text-video generation~\cite{wu2023tune} etc. 
On the other hand, DRL can also be applied in advanced models such as diffusion models, demonstrating the wide applicability of DRL.
As such, we will briefly introduce the definition of diffusion model and elaborate on several diffusion-based DRL methods.

Given that diffusion model tends to heavily involve time step, in this section we employ subscripts to indicate different time steps and adopt superscripts to denote different factors.
Formally, diffusion model defines a forward process that adds the noise to the real data $\mathbf{x}_{0}$:

{
\small
\begin{align}
x_{t} = \alpha_t x_{0} +  \sigma_t \boldsymbol{\epsilon}, \ \boldsymbol{\epsilon} \sim \mathcal{N}(\mathbf{0}, \mathbf{I}), 
\end{align}
}

\noindent where $\alpha_t$ and $\sigma_t$ are functions of $t$ and the signal-to-noise-ratio $\alpha_t / \sigma_t$ is strictly decreasing. 
Diffusion models train a neural network ${\boldsymbol{\epsilon}_{\theta}}$ to predict the noise of noisy variable $\mathbf{x}_{t}$, whose training objective is:

{\small
\begin{align}
\mathbf{E}_{x_{0}, \boldsymbol{\epsilon}, t} 
\bigg[\big\|\boldsymbol{\epsilon}_{\theta}\left(\alpha_{t} x_{0}+\sigma_{t} \boldsymbol{\epsilon},t \right)-\boldsymbol{\epsilon}\big\|_{2}^{2}\bigg]  .
\end{align}
}

In the reverse process, diffusion model generates samples by gradually denoising from a Gaussian noise at time step $T$, i.e., $x_T \sim \mathcal{N}(\mathbf{0}, \mathbf{I})$, to real data, i.e., $x_0$, via $p_\theta(x_{t-1}|x_t)=\mathcal{N}(x_t;\mu_\theta(x_t,t),\sigma_tI)$, where $\mu_\theta(x_t,t)$ is determined by the diffusion network $\boldsymbol{\epsilon}_{\theta}$ with certain sampling trajectories~\cite{ho2020denoising, liupseudo, lu2022dpm}.

Diffusion model can also benefit from the strength of DRL. Yang et al.~\cite{yang2023disdiff} propose an unsupervised disentanglement framework DisDiff whose network structure is shown in Figure~\ref{fig:disdiff}. They learn a disentangled representation and a disentangled gradient field for each generative factor. Specifically, they assign a separate encoder $E_{\phi}^{i}$ for each factor $f^i$ to extract the disentangled representations $z^i$, i.e., $\{z^{1},z^{2},\ldots,z^{N}\} = \{E_{\phi}^{1}(x_{0}),E_{\phi}^{2}(x_{0}),\ldots,E_{\phi}^{N}(x_{0})\}$. Motivated by class-guided~\cite{dhariwal2021diffusion} sampling of diffusion models, they employ a decoder $G_{\psi}(x_{t},z,t)$ to estimate the score function, i.e., gradient field $\nabla_{x_{t}}\log p(z^{i}|x_{t})$. Then the sampling process can be written as follows,

{\small
\begin{align}
p_\theta(x_{t-1}|x_t) = \mathcal{N}(x_{t-1};\mu_{\theta}(x_{t},t)+\sigma_{t}\sum_{i}\nabla_{x_{t}}\log p(z^{i}|x_{t}),\sigma_{t}).
\label{eq:disdiff:1}
\end{align}
}

\noindent The next critical point is to ensure the disentanglement among these representations $z^i$.  The authors propose the following disentanglement loss that minimizes the upper bound for the mutual information between $z^i$ and $z^j,  \text{where} \ i \neq j$:

{\small
\begin{align}
\mathcal{L}_{dis} = \operatorname*{min}_{i,j,x_{0},\hat{x}_{0}^{j}}\|\hat{z}^{i|j}-\hat{z}^{i}\|-\|\hat{z}^{i|j}-z^{i}\|+\|\hat{z}^{j|j}-z^{j}\|,
\label{eq:disdiff:2}
\end{align}
}

\noindent where $\hat{x}_0^j$ is conditioned on $z^j$ through the sampling process $p_\theta(x_{t-1}|x_t,z^j)$, $\hat{z}^{i}=E_{\phi}^{i}(\hat{x}_{0})$, and $\hat{z}^{i|j}=E_{\phi}^{i}(\hat{x}_{0}^{j})$. Note that $\hat{x}_0$ denotes the sampled data, while $x_0$ denotes the groundtruth data.

\begin{figure}
    \centering
    \includegraphics[width=0.85\linewidth]{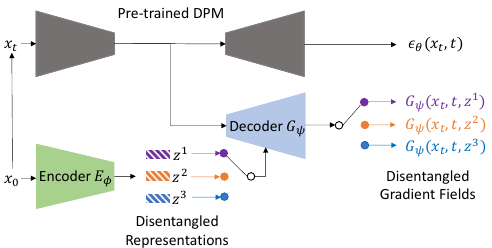}
    \caption{The network structure of DisDiff, figure from~\cite{yang2023disdiff}.}
    \label{fig:disdiff}
\end{figure}

Chen et al.~\cite{chen2023disenbooth}
propose a disentangled stable-diffusion~\cite{rombach2022high} fine-tuning framework, DisenBooth, for subject-driven text-to-image generation by disentangling the features into the identity-preserved and the identity-irrelevant parts. Specifically, they adopt a CLIP text encoder to extract textual identity-preserved embedding for the text, and employ an adapter to extract visual identity-irrelevant embedding for each image (e.g., the background). To ensure the disentanglement, they also use several specifically designed loss functions, including weak denoising loss and contrastive loss. To further enhance the disentanglement, the subsequent work~\cite{chen2024disendreamer} introduces sample-specific mask and sample-specific denoising loss.

Chen et al.~\cite{chen2023videodreamer} further propose a disentangled fine-tuning framework, VideoDreamer, for multi-subject-driven video generation. They mix all the input subjects in a single picture, and then utilize three disentangled features to represent three independent components, i.e., i) image-specific information such as background and pose, ii) stitch condition caused by mixture, and iii) mixed subject-identity information.
}

\subsubsection{Disentangled Latent Directions in Pretrained Generative Models}

\red{Most GAN-based and VAE-based DRL methods conform to the paradigm of training from scratch with certain disentanglement loss functions. However, recent works~\cite{shen2021closed,khrulkov2021disentangled,voynov2020unsupervised} have shown semantically meaningful variations when traversing along certain directions in the latent space of pretrained generative models. The phenomenon indicates that there exist certain properties of disentanglement in the latent space of the pretrained generator. Each meaningful and interpretable latent direction aligns with one generative factor.
Discovering the disentangled latent directions is a more efficient way to achieve DRL than conventional methods, which can leverage the power of pretrained models and also save resources. Pretrained models play a very important role, especially in the era of big models. Efficiently equipping big models with disentanglement ability by discovering disentangled latent directions may serve as a promising path. In this section we will delve into several representative trials in this path, from GANs to diffusion models.

Voynov et al.~\cite{voynov2020unsupervised} propose the first unsupervised framework for the discovery of disentangled latent directions in the pretrained GAN latent space. Specifically, they learn a matrix $A\in\mathbb{R}^{d\times K}$ where $d$ denotes the dimension of latent space and $K$ is the total number of generative factors. The $k$-th column represents the direction vector of the $k$-th factor. The direction vectors can be added to a latent code $z$ to transform corresponding generative factors in the image space. To learn $A$, they obtain image pairs in the form of $\left(G(z),G(z+A(\varepsilon e_k))\right)$, where $z\sim\mathcal{N}(0,I)$, G denotes the generator, $e_k$ denotes an axis-aligned unit vector, and $\varepsilon$ denotes a shift scalar. A reconstructor R is employed to predict the index $k$ and the shift $\varepsilon$ given the generated pairs $\left(G(z),G(z+A(\varepsilon e_k))\right)$. This loss function is shown as follows,

{\small
\begin{align}
\min _{A, R} \underset{z, k, \varepsilon}{\mathbb{E}} L(A, R)=\min _{A, R} \underset{z, k, \varepsilon}{\mathbb{E}}\left[L_{c l}(k, \widehat{k})+\lambda L_{r}(\varepsilon, \widehat{\varepsilon})\right] ,
\label{eq:voybov }
\end{align}
}

\noindent where $\hat{k}, \hat{\varepsilon} = R\Big(G(z),G\big(z+A(\varepsilon e_k)\big)\Big)$. Minimizing this loss will force $A$ to form disentangled direction vectors that are easier to be distinguished by the reconstructor R.

Ren et al.~\cite{ren2021learning} propose an unsupervised framework Disco with contrastive learning for the discovery of latent directions. Similar to Voynov et al.~\cite{voynov2020unsupervised}, they also utilize a learnable matrix $A\in\mathbb{R}^{d\times K}$ to represent latent directions, with each column representing a candidate direction. Besides, they employ an encoder $E$ to explicitly extract disentangled representations. Note that ``disentangled representations'' here are not in the original latent space of the pretrained generator $G$, but in a separate space extracted by the encoder. Therefore, the discovered disentangled latent directions can be used to conduct disentangled controllable generation, while the extracted disentangled representations can be applied in downstream tasks. The authors adopt contrastive learning to jointly optimize the latent directions $A$ and the encoder $E$. Specifically, they first construct a variation space by $\boldsymbol{\upsilon}(z,k,\varepsilon)=\big|E\big(G(z+A(\varepsilon e_k))\big)-E(G(z))\big|$, where $k$ denotes the direction index and $\varepsilon$ denotes the shift. Then the contrastive loss is employed to pull together the variation samples with the same $k$ and push away the ones from different $k$.

Kwon et al.~\cite{kwon2022diffusion} propose a framework called Asyrp for the image-editing task, by discovering the semantic directions in the space of the deepest feature maps of a pretrained diffusion model. Specifically, they align the direction ${\Delta h}_i$ to the $i$-th attribute such as ``smiling'', where $h$ is the deepest feature maps of the pretrained UNet. Then they can edit the attribute by shifting along the direction, i.e., $\Tilde{h} = h+ \alpha {\Delta h}_i$. Then this shift will change the UNet output, i.e., the predicted noise $\epsilon_{\theta}$, and finally change the $i$-th attribute of generated images. They employ a CLIP-based directional loss with cosine distance to optimize ${\Delta h}_i$.

Besides, DisDiff~\cite{yang2023disdiff} introduced in Section~\ref{sec: diffusion methods} can also be regarded as a direction discovery method, as it obtains disentangled latent directions $\nabla_{x_{t}}\log p(z^{i}|x_{t})$ in the gradient fields.
}




\subsubsection{Comparison}
\red{As we mention in previous sections, VAE-based methods usually suffer from the trade-off between disentanglement performance (i.e., explanation ability) and generative ability (i.e., reconstruction performance). Although GAN-based models usually have a remarkable generative ability, they lack the reversibility property, making them less flexible than VAEs. It seems that the more recent generative models, e.g., diffusion models, are able to combine the advantages of GANs and VAEs. Diffusion models not only possess powerful generative abilities, but also naturally have a friendly reversibility property, demonstrating huge potential to become the mainstream in future DRL research.}

\subsection{Representation Structure}
We in detail discuss DRL regarding its representation structure in terms of i) dimension-wise v.s. vector-wise, and ii) flat v.s. hierarchical.

\subsubsection{Dimension-wise DRL v.s. Vector-wise DRL}
\label{sec:dim vs. vec}
\begin{figure}[htp]
    \centering
    \includegraphics[width=1.0\linewidth]{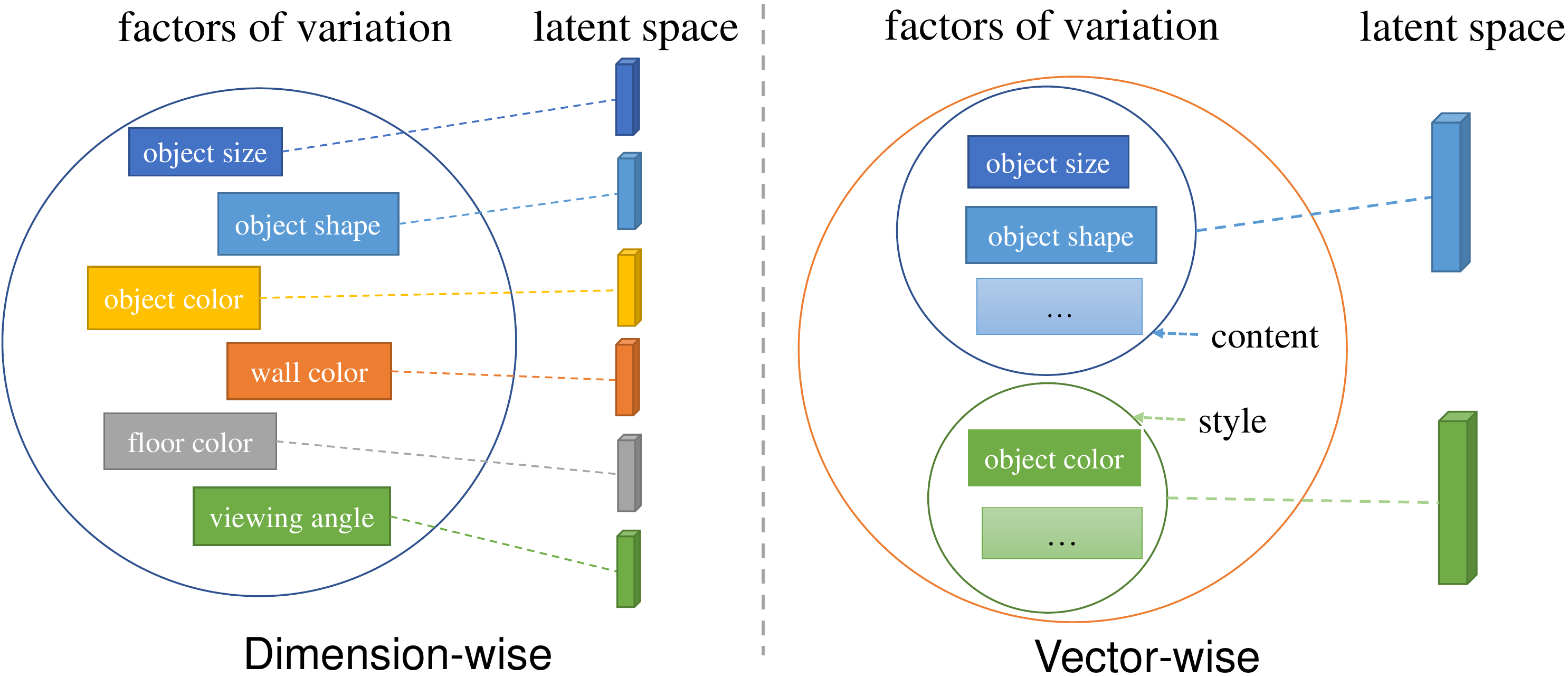}
    \caption{The illustration of the comparison between dimension-wise and vector-wise DRL.}
    \label{fig: dim vs. vec}
\end{figure}

\red{
Based on the structure of disentangled representations, we can categorize DRL methods into two groups, i.e., dimension-wise and vector-wise methods. 
Let $k_1$ and $k_2$ denote the total number of generative factors for dimension-wise and vector-wise methods.
Dimension-wise methods adopt a set of disentangled dimensions $\textbf{z}=\{z_i,i=0,..k_1-1\}$, where $\textbf{z}$ is the whole representation and each single dimension $z_i$ (a 1-dimension scalar) represents one fine-grained generative factor.
In contrast, vector-wise methods employ $k_2$ disentangled vectors $\textbf{z}_i,i=0,..,k_2-1$ to represent $k_2$ coarse-grained factors, where the dimension of each vector $\textbf{z}_i$ equals to or is larger than 2.
The comparisons of dimension-wise and vector-wise methods are shown in Figure~\ref{fig: dim vs. vec} and Table~\ref{tab:dimension-vs-vector}. Dimension-wise methods are always experimented on synthetic and simple datasets, while vector-wise methods are always used in real-world tasks such as identity swapping, image classification, subject-driven generation, and video understanding etc. Synthetic and simple datasets usually have fine-grained latent factors, leading to the applicability of dimension-wise disentanglement. In contrast, for real-world datasets and applications, we usually concentrate on coarse-grained factors (e.g., identity and pose), making it more suitable for vector-wise disentanglement. Dimension-wise methods are mostly early theoretical explorations of DRL, and usually rely on certain model architectures, e.g., VAE or GAN. Vector-wise methods are more flexible, for example, we can design task-specific encoders to extract disentangled vectors and design appropriate loss functions to ensure disentanglement.

The fundamental difference between dimension-wise and vector-wise methods lies in the information capacity determined by the number of dimensions to represent a factor. Specifically, dimension-wise methods usually allocate 1-dimension scalar to represent a generative factor, while vector-wise methods employ more dimensions in the form of vectors to represent generative factors. With more dimensions, vector-wise representations naturally are more powerful than dimension-wise representations to represent complex information. Therefore, we believe this is the main reason why vector-wise DRL is suitable for coarse-grained factors capturing more information while dimension-wise DRL is suitable for fine-grained factors capturing relatively less information. 
However, we note that granularity is sometimes a relative concept. For example,  on the one hand ``object size'' in simple synthetic datasets is usually a fine-grained factor, which is enough to be represented via a single dimension scalar. On the other hand, ``object size'' can be a coarse-grained concept in complex real-world scenes by decomposing “object size” into more fine-grained concepts such “width” and “height”, which needs vectors with more dimensions to represent. Therefore, we remark that in practice the choice of dimension-wise or vector-wise heavily relies on the complexity of target scenes and tasks. 
}
In this section, we will discuss a series of representatives of dimension-wise and vector-wise DRL methods.

\begin{table*}[th]
\small
\centering
\caption{The comparisons of dimension-wise and vector-wise methods.}
\label{tab:dimension-vs-vector}
\begin{tabular}{|m{1.5cm}<{\centering}|m{3cm}<{\centering}|m{5cm}<{\centering}|m{4cm}<{\centering}|m{3cm}<{\centering}|} 
\hline
   Methods    & Dimension of Each Latent Factor & Representative Works       & Semantic Alignment   & Applicability    \\ 
\hline
Vector-wise   & multiple   & MAP-IVR~\cite{liu2021activity}, DRNET~\cite{denton2017unsupervised}, DR-GAN~\cite{tran2017disentangled}, DRANet~\cite{Lee_2021_CVPR}, Lee et al.~\cite{Lee_2018_ECCV}, Liu et al.~\cite{Liu_2021_CVPR}, Singh et al.~\cite{singh2019finegan}

& each latent variable aligns to one coarse-grained semantic meaning & real scenes    \\ 
\hline
Dimension-wise & one    &

VAE-based methods, InfoGAN~\cite{chen2016infogan}, IB-GAN~\cite{jeon2021ib}, Zhu et al.~\cite{zhu2018visual}, InfoGAN-CR~\cite{lin2019infogan}, PS-SC GAN~\cite{zhu2021and}, Wei et al.~\cite{wei2021orthogonal}, DNA-GAN~\cite{xiao2017dna}

& each dimension aligns to one fine-grained semantic meaning  & synthetic and simple datasets  \\
\hline
\end{tabular}
\end{table*}

\noindent \textbf{Dimension-wise Methods.} A typical architecture that can be used to achieve dimension-wise disentanglement is the VAE-based method, on which we have spent a lot of words in Section~\ref{method:VAE}. Besides, GAN-based methods are another important paradigm that can achieve dimension-wise DRL, as discussed in Section~\ref{sec:gan-based method}. Next, we will focus more on various vector-wise methods.

\noindent \textbf{Vector-wise Methods.} Besides dimension-wise DRL, GAN-based models can also be used to achieve vector-wise DRL on more real-world tasks. Tran et al.~\cite{tran2017disentangled} propose DR-GAN for pose-invariant face recognition. They use two vectors to represent identity and pose respectively. Specifically, they explicitly set a one-hot latent vector to represent the pose and use an encoder to extract the identity vector from input images. Two Discriminators
for pose and identity respectively are used to ensure that the latent vectors can align with the corresponding semantics (i.e., generative factors). Finally, the learned identity vector can be used to conduct pose-invariant face recognition or synthesize identity-preserving faces.

Liu et al.~\cite{liu2021activity} propose a disentangled framework MAP-IVR for activity image-to-video retrieval, which separates video representation into appearance and motion parts. Specifically, they use two video encoders to respectively extract the video motion feature $m^v$ and video appearance feature $a^v$ from the video feature $\bar{v}$. They also use an image encoder to extract the appearance feature of the reference image.
They design several objectives to ensure the disentanglement:

{\small
\begin{align}
\mathcal{L}_{orth}=\cos(m^v,a^v),
\label{eq:map-ivr:1}
\end{align}
}

{\small
\begin{align}
\mathcal{L}_{class}=-\log(p(a^v)_y)-\log(p(a^u)_y),
\label{eq:map-ivr:2}
\end{align}
}

{\small
\begin{align}
\mathcal{L}_{re}=\|\bar{v}-\hat{v}\|_2^2.
\label{eq:map-ivr:3}
\end{align}
}

\noindent $\mathcal{L}_{orth}$ is a cosine distance loss that facilitates the orthogonality between the video motion and appearance feature. $\mathcal{L}_{class}$ leverages an activity classifier $p$ to ensure that the video appearance feature $a^v$ and the image appearance feature $a^u$ can both capture the activity information. The reconstruction loss $\mathcal{L}_{re}$ also helps the disentanglement, where $\bar{v}$ is the original video feature and $\hat{v}$ is reconstructed by combining $a^u$ and the motion information from $m^v$. The disentangled features can be used to accomplish better activity image-to-video retrieval by translating the image reference to a video reference.

Denton et al. propose an autoencoder-based model DRNET~\cite{denton2017unsupervised} that disentangles each video frame into a time-invariant (content) and a time-varying (pose) component. They use two encoders to extract the content feature and the pose feature respectively. Let $h_c^t$ and $h_p^t$ denote the content feature and the pose feature of the $i$-th frame. The disentanglement objectives are:

{\small
\begin{align}
\mathcal{L}_{re}(D)=||D(h_c^t,h_p^{t+k})-x^{t+k}||_2^2,
\label{eq:drnet:1}
\end{align}
}

{\small
\begin{align}
\mathcal{L}_{sim}(E_c)=||E_c(x^t)-E_c(x^{t+k})||_2^2.
\label{eq:drnet:2}
\end{align}
}

\noindent $\mathcal{L}_{re}(D)$ is the reconstruction loss that aims to ensure combining $h_c^t$ and $h_p^{t+k}$ can reconstruct the frame $x^{t+k}$. $\mathcal{L}_{sim}$ means $h^c_t=E_c(x^t)$ should be invariant across t. The two losses expect $h_c$ to capture the time-invariant content and $h_p$ to capture the time-varying pose. Besides, they also use an adversarial loss to help $h_p$ not carry information about the content. The disentangled representations can be applied in future frames prediction or classification tasks.

Cheng et al.~\cite{cheng2021disentangled} propose DFR that leverages disentangled features to achieve few-shot image classification. They utilize two encoders $E_{cls}$ and $E_{var}$ to extract class-specific and class-irrelevant features, respectively. The disentangled objectives are:

{\small
\begin{align}
\mathcal{L}_{dis}=-\sum_{i=1}^P\left(l_i\cdot log(s_i)+(1-l_i)\cdot log(1-s_i)\right),
\label{eq:DFR:1}
\end{align}
}

{\small
\begin{align}
\mathcal{L}_{cls} =-\sum_{i=1}y_i\log P\left(\hat{y}_i=y_i\mid\mathcal{T}_{FS}\right), 
\label{eq:DFR:2}
\end{align}
}



\noindent $\mathcal{L}_{dis}$ is a discriminative loss that removes the class-specific information in the class-irrelevant feature. $s_i=r_\varphi\left(E_{var}(x_{i1}), E_{var}(x_{i2})\right)$, where $r_\varphi$ is the discriminator which outputs the probability that the pair $x_{i1}$ and $x_{i2}$ are from the same class. They employ a gradient reversal layer to encourage $E_{var}$ to remove the class-specific information. $\mathcal{L}_{cls}$ is the cross-entropy loss for classification loss, where the prediction $\hat{y}_i$ is obtained on the basis of the class-specific feature $E_{cls}(x_i)$. $\mathcal{L}_{cls}$ encourages $E_{cls}$ to capture class-specific information. Besides, they employ a reconstruction loss and a translation loss to further promote disentanglement.

Lee et al. propose a disentangled cross-domain adaptation framework DRANet~\cite{Lee_2021_CVPR}. They use one encoder to extract the content feature while obtaining the style feature by subtracting the content feature from the original feature. They also design several losses, e.g., a perceptual loss to enhance disentanglement. Domain adaption can be achieved by combining the content feature with the style feature of the target domain. Gao et al. propose a disentangled identity-swapping framework InfoSwap~\cite{Gao_2021_CVPR}, which disentangles identity-relevant and identity-irrelevant. They achieve disentanglement by optimizing a loss objective based on the information bottleneck theory. Identity-swapping can be achieved by combining the identity-relevant feature with the target identity-irrelevant feature.

\noindent \textbf{Discussion.} We have introduced a variety of dimension-wise and vector-wise disentangled representation learning methods. 
Dimension-wise DRL methods use a single dimension (or several dimensions) to represent one fine-grained generative factor, while vector-wise DRL methods use a single vector to represent one coarse-grained generative factor.
A common key point of them is how to enforce disentanglement by designing certain loss objectives, e.g., various regularizers or specifically-designed supervised signals. We will discuss in depth how to design loss objectives for DRL tasks in Sec \ref{sec:designs}. As for how to determine which kind of structure (i.e., dimension-wise or vector-wise) to use in different tasks, it depends on the number and the granularity of the generative factors we hope to take into account. For example, in many real-world applications, we only need to consider two or several coarse-grained factors. On the other hand, for some specific generative tasks on simple datasets, we need to consider multiple fine-grained factors. The dimension-wise methods are mostly early theoretical exploration for DRL in simple scenes, while in recent years, researchers focus more on how to incorporate vector-wise DRL to tackle real-world applications. It might be a trend to explore the power of vector-wise DRL in various realistic tasks.





\subsubsection{Flat DRL vs. Hierarchical DRL}

The aforementioned DRL methods hold an assumption that the architecture of generative factors is flat, i.e., all the factors are parallel and at the same abstraction level. For example, as for dimension-wise DRL, $\beta$-VAE~\cite{higgins2016beta} disentangles face rotation, smile, skin color, fringe, etc. on CelebA dataset. InfoGAN~\cite{chen2016infogan} disentangles azimuth, elevation, lighting, etc. on 3D Faces dataset. As for vector-wise DRL, DR-GAN~\cite{tran2017disentangled} disentangles face identity and pose. MAP-IVR~\cite{liu2021activity} disentangles motion and appearance features for video. DisenBooth~\cite{chen2023disenbooth} disentangles the identity-preserved and the identity-irrelevant features. In summary, there doesn't exist a hierarchical structure among these disentangled factors.

\begin{figure}[htp]
    \centering
    \includegraphics[width=0.72\linewidth]{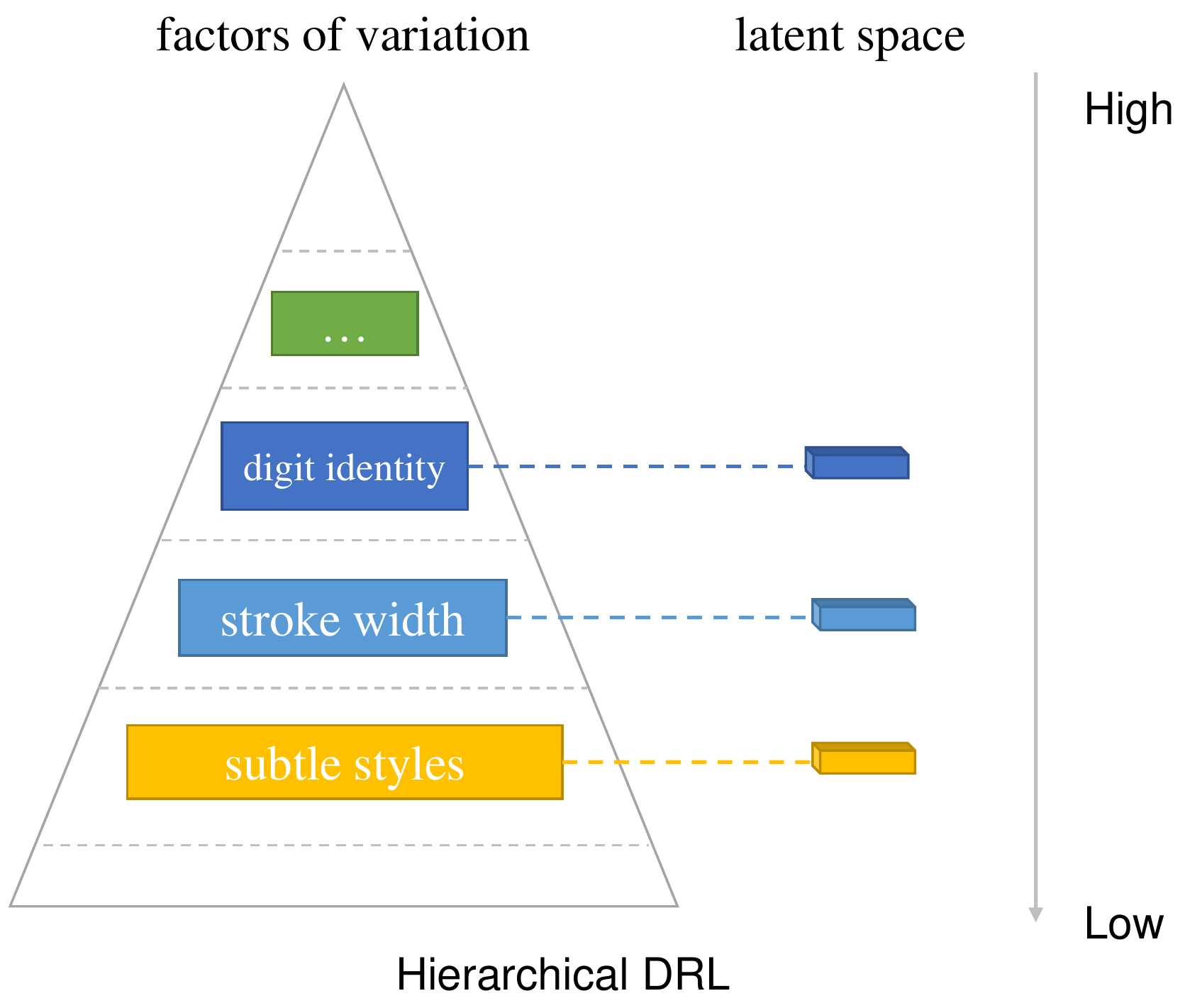}
    \caption{The illustration of hierarchical DRL. There exists a hierarchical structure among generative factors, i.e., the factors belong to different abstraction levels, resembling a pyramid. }
    \label{fig: hierarchical DRL}
\end{figure}

However, in practice, generative processes might naturally involve hierarchical structures~\cite{ross2021benchmarks,li2020progressive} where the factors of variation have different levels of semantic abstraction, either dependent~\cite{ross2021benchmarks} or independent~\cite{li2020progressive} across levels. 
For example, the factor controlling \textit{gender} has a higher level of abstraction than the independent factor controlling \textit{eye-shadow} on CelebA dataset~\cite{li2020progressive}, 
while there exist dependencies between factors controlling \textit{shape} (higher level) and \textit{phase} (lower level) on Spaceshapes dataset~\cite{ross2021benchmarks}, e.g., the dimension of ``phase" is active only when the object shape equals to ``moon". 
To capture these hierarchical structures, a series of works have been proposed to achieve hierarchical disentanglement. Figure~\ref{fig: hierarchical DRL} demonstrates the paradigm of hierarchical DRL.

\begin{figure}[htp]
    \centering
    \includegraphics[width=0.90\linewidth]{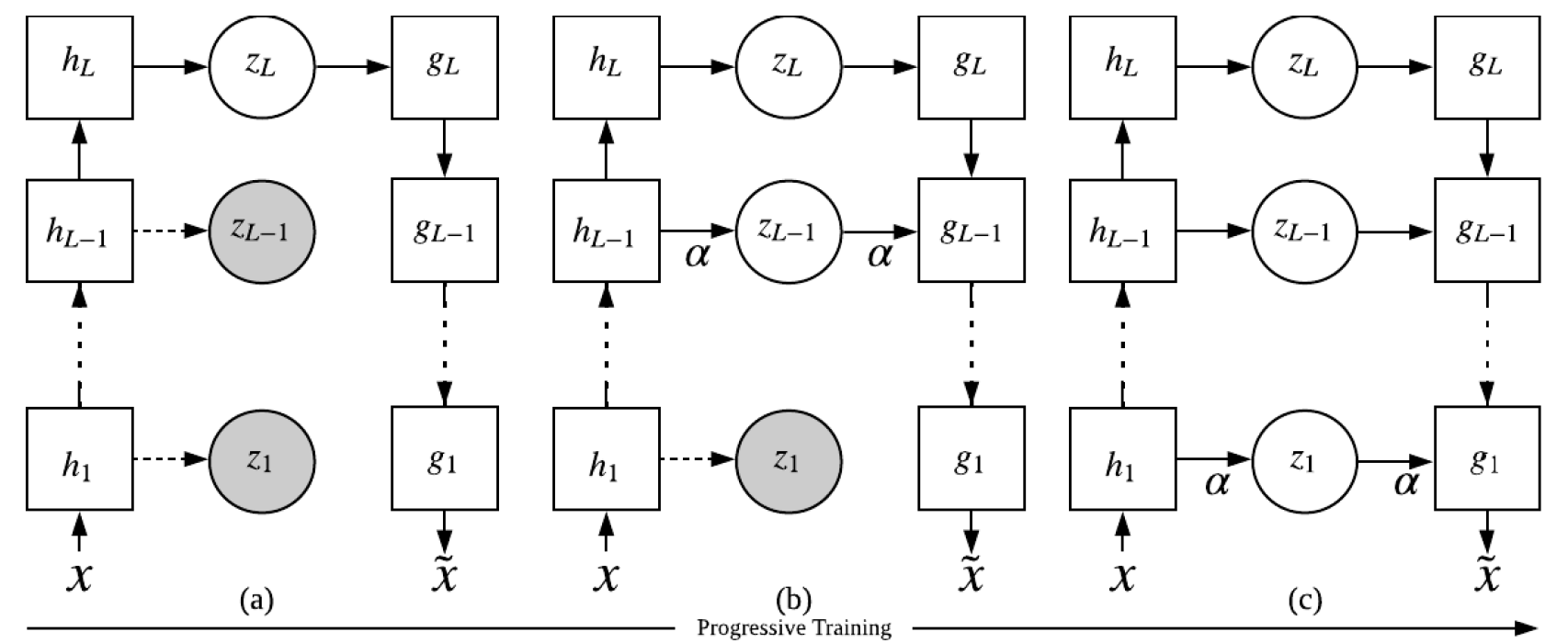}
    \caption{The architecture of the hierarchical framework proposed by Li et al.~\cite{li2020progressive}. The figure is from the original paper.}
    \label{fig:hierarchical vae}
\end{figure}

Li et al.~\cite{li2020progressive} propose a VAE-based model which learns hierarchical disentangled representations through formulating the hierarchical generative probability model in Eq.~\eqref{eq:hierarchical1},

{\small
\begin{align}
p(\mathbf{x}, \mathbf{z})=p\left(\mathbf{x} \mid \mathbf{z}_{1}, \mathbf{z}_{2}, \ldots, \mathbf{z}_{L}\right) \prod_{l=1}^{L} p\left(\mathbf{z}_{l}\right),
\label{eq:hierarchical1}
\end{align}
}

\noindent where $\mathbf{z}_l$ denotes the latent representation of the $l$-th level abstraction, and a larger value of $l$ indicates a higher level of abstraction. 
The authors estimate the level of abstraction with the network depth, i.e., the deeper network layer is responsible for outputting representations with higher abstraction level. 
It is worth noting that Eq.\eqref{eq:hierarchical1} assumes that there is no dependency among latent representations with different abstraction levels. 
In other words, each latent representation tends to capture the factors that belong to a single abstraction level, which will not be covered in other levels. 
The corresponding inference model is formulated in Eq.\eqref{eq:hierarchical2} as follows,

{\small
\begin{align}
q\left(\mathbf{z}_{1}, \mathbf{z}_{2}, \ldots, \mathbf{z}_{L} \mid \mathbf{x}\right)=\prod_{l=1}^{L} q\big(\mathbf{z}_{l} \mid \mathbf{h}_{l}(\mathbf{x})\big),
\label{eq:hierarchical2}
\end{align}
}

\noindent where $\mathbf{h}_{l}(\mathbf{x})$ represents the abstraction of $l$-th level. In the training stage, the authors design a progressive strategy of learning representations from high to low abstraction levels with modified ELBO objectives. 
The hierarchical progressive learning is shown in Figure~\ref{fig:hierarchical vae}, where $h_i$ and $g_i$ are a set of encoders and decoders at different abstraction levels. The framework can disentangle digit identity, stroke width, and subtle digit styles on MNIST dataset, from high to low abstraction levels. It can also disentangle gender, smile, wavy-hair, and eye-shadow on CelebA.

Tong et al.~\cite{tong2019hierarchical} propose to learn a set of hierarchical disentangled representations $\mathbf{z}=\left\{\mathbf{z}_{l}^{i}\right\}_{i=1}^{c_{l}}$, where $\mathbf{z}_{l}^{i}$ is the $i$-th latent variable of the $l$-th layer in the hierarchical structure and $c_l$ is the total number of latent variables of the $l$-th layer. To ensure disentanglement at each hierarchical level, they design a loss function shown in Eq.\eqref{eq:hierarchical3},

{\small
\begin{align}
\mathcal{L}_{\text {disentangle }}=\sum_{l} \frac{2}{c_{l}\left(c_{l}-1\right)} \sum_{i \neq j}^{c_{l}} \mathrm{dCov}^{2}\left(\mathbf{z}_{l}^{i}, \mathbf{z}_{l}^{j}\right),
\label{eq:hierarchical3}
\end{align}
}
\noindent where $\mathrm{dCov}^{2}(\cdot, \cdot)$ denotes the distance covariance.

Singh et al.~\cite{singh2019finegan} propose an unsupervised hierarchical disentanglement framework FineGAN for fine-grained object generation. 
They design three latent representations for different hierarchical levels, i.e., background code $\mathbf{b}$, parent code $\mathbf{p}$ and child code $\mathbf{c}$, which represent background, 
object shape and object appearance respectively. 
Background is the lowest level, followed by shape and appearance. 
In the generation process, FineGAN first generates a realistic background image by taking $\mathbf{b}$ and noise $\mathbf{z}$ as input. 
Then it generates the shape and stitches it on top of the background image through taking $\mathbf{p}$ and noise $\mathbf{z}$ as input. 
Finally, by taking $\mathbf{c}$ as input conditioned on $\mathbf{p}$, the model fills in the shape (parent) outline with appearance (child) details. 
The authors further employ information theory (similar to InfoGAN) to disentangle the parent (shape) and child (appearance), 
and use an adversarial loss together with an auxiliary background classification loss to constrain the background generation.

Li et al.~\cite{li2021image} propose a hierarchical disentanglement framework for image-to-image translation. 
They manually organize the labels into a hierarchical tree structure from root to leaves and from high to low level of abstraction, for example, \textit{tags} (e.g., glasses), \textit{attributes} (e.g., with or without), \textit{styles} (e.g., myopic glasses, sunglasses). It is worth noting that the tree hierarchical structure indicates that the child nodes depend on their parents. 
The authors train a translator module to deal with tags and train an encoder to extract style features.

Ross et al.~\cite{ross2021benchmarks} propose a hierarchical disentanglement framework, which assumes that a group of dimensions may only be active in some cases. 
Specifically, they organize generative factors as a hierarchical structure (e.g., tree) such that whether a child node can be active depends on the value of its parent node. 
Take the Spaceshapes dataset as an example, the dimension representing \textit{phase} will only be active when the value of its parent \textit{shape} equals to ``moon". 
They design an algorithm named MIMOSA to
train an autoencoder to learn the hierarchical disentangled representations.


\noindent \textbf{Discussion.} In summary, we can choose the flat or hierarchical DRL methods, depending on whether the generative factors have a hierarchical structure. Although flat DRL methods might also have the potential to disentangle factors from different levels of abstraction, the hierarchical DRL methods with particular designs for the hierarchical structure perform much better. In specific applications, we can consider if there is an underlying hierarchical structure that we can leverage to facilitate disentanglement.

\subsection{Supervision Signal}
\label{unsupervised vs. supervised}

In this section, we revisit the DRL methods from the perspective of learning schemes, i.e., unsupervised vs. supervised DRL, which serve as a fundamental issue attracting a lot of attention as well as a crucial problem in achieving disentanglement.

\subsubsection{Unsupervised Methods}
Major schemes in DRL methods mainly stem from unsupervised learning, particularly pursuing automated discovery of interpretable factorized latent representations in early studies.
The original VAE~\cite{kingma2013auto} model demonstrates the possibility of learning latent space in unsupervised manner using Bayesian inference.
A class of adversarial generative network models represented by InfoGAN~\cite{chen2016infogan} strive to learn explainable representations in an unsupervised manner.
Additionally, research grounded in information theory also introduces methods such as mutual information estimation~\cite{hjelm2018learning} and DeepVIB~\cite{alemi2019deep} to disentangle underlying information and achieve better robustness and generalization ability.

As a primitive stage of development in DRL, unsupervised learning paradigm is built as the original vision for most researchers, which represents a class of intuitive and effective implementation methods of DRL.

\subsubsection{Supervised Methods}
Locatello et al.~\cite{locatello2019challenging} prove that "pure unsupervised DRL is theoretically impossible without inductive bias on methods and datasets" recently. In other words, disentanglement itself does not occur naturally, which breaks the situation that researchers have been focusing on "unsupervised disentanglement".
Locatello et al.~\cite{Locatello2020Disentangling} later propose that using some of the labeled data for training is beneficial both in terms of disentanglement and downstream performance.

DC-IGN~\cite{kulkarni2015deep} restricts only one factor to be variant and others to be invariant in each mini-batch. One dimension of latent representation $\mathbf{z}$ is chosen as $z_{train}$ which is trained to explain all the variances within the batch
and through supervision, thus aligns to the selected variant factor. ML-VAE~\cite{bouchacourt2018multi} divides samples into groups according to one selected factor $f_s$, where samples in each group share the same value of $f_s$. This setting is more applicable for some applications such as image-to-image translation, where images in each group share the same label as well as the same posterior of latent variables with respect to $f_s$, which depends on all the samples in the group. 
While as for other factors except $f_s$, the posterior may be dependent on each individual sample.

Besides, Bengio et al.~\cite{Bengio2020A} claim that adaptation speed can evaluate how well a model fits the underlying causal structure from the view of causal inference, and then propose exploiting a meta-learning objective to learn well-represented, disentangled and structured causal representations given an unknown mixture of causal variables.

\begin{figure}[htp]
    \centering
    \includegraphics[width=0.85\linewidth]{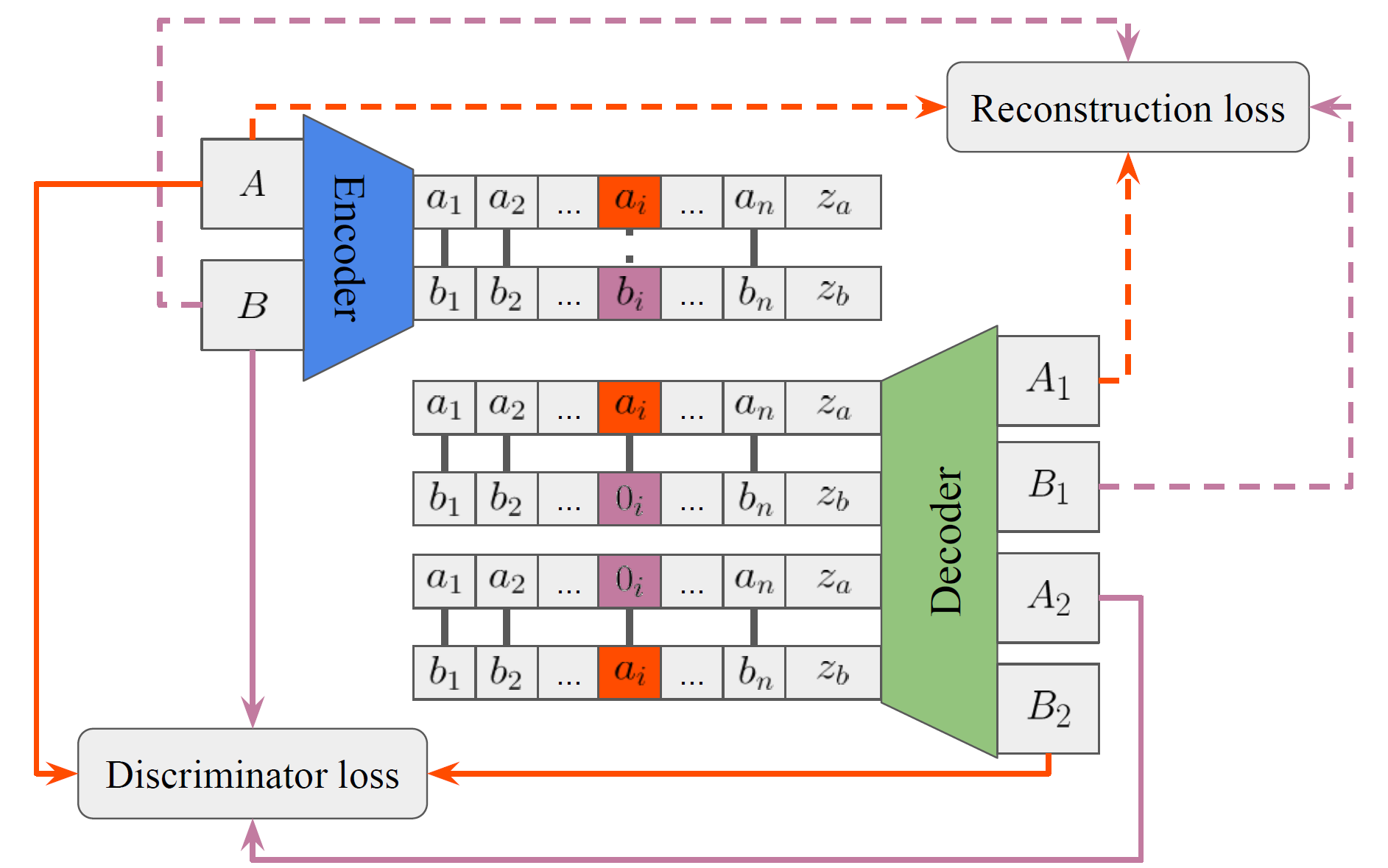}
    \caption{The architecture of DNA-GAN, figure from~\cite{xiao2017dna}.}
    \label{fig:dna-gan}
\end{figure}

Xiao et al.~\cite{xiao2017dna} propose DNA-GAN, a supervised model whose training procedure similar to gene swap. In concrete, DR-GAN takes a pair of multi-labeled images $I_a$ and $I_b$ with different labels as the input of the encoder. 
After obtaining the original representations $\mathbf{a}$ and $\mathbf{b}$ of $I_a$ and $I_b$ through an encoder, the swapped representations $\mathbf{a}^\prime$ and $\mathbf{b}^\prime$ are constructed by swapping the value of a particular dimension in the attribute-relevant part of the original representations. After decoding, the reconstruction and the adversarial loss are applied to ensure that each dimension of attribute-relevant representations can align with the corresponding labels.The architecture is shown in Figure~\ref{fig:dna-gan}

Moreover, the guidance from reconstruction loss and task loss can also be regarded as supervised signals, which are widely used in real-world applications. 
We will elaborately discuss this in the section ``Designs of DRL " (Sec.~\ref{sec:designs}).

\subsubsection{Weakly Supervised Methods}
\red{
Supervised DRL methods hold the assumption that the target dataset is semantically clear and well-structured to be disentangled into 
explanatory, independent, and recoverable generative factors~\cite{xiang2021disunknown}. 
However, in some cases there exist intractable factors which are unclear or difficult to annotate, where these factors are usually regarded as noises unrelated to the target task.
}

\red{
Xiang et al.~\cite{xiang2021disunknown} propose a weakly-supervised DRL framework, DisUnknown, with the setting of $N-1$ factors labeled and $1$ factor unknown out of totally $N$ factors. As such, all the intractable factors or task-irrelevant factors can be covered in a single unknown factor. The DisUnknown model is a two-stage method including i) unknown factor distillation and ii) multi-conditional generation, where the first stage extracts the unknown factor by adversarial training and the second stage embeds all labeled factors for reconstruction.
They use a set of discriminative classifiers that predict the probability distribution of factor labels to enforce disentanglement, similar to the idea of InfoGAN~\cite{chen2016infogan}.
}

\subsubsection{The Identifiability of DRL}

One of the most significant concerns of disentangled representation learning is the identifiability~\cite{locatello2019challenging, Yang_2021_CVPR, khemakhem2020variational}. It is mainly discussed in the context of unsupervised DRL, which focuses on the feasibility of unsupervised DRL. The identifiability indicates whether we can distinguish the disentangled model that we expect to obtain from other entangled ones. Locatello et al.~\cite{locatello2019challenging} claim that it is impossible to identify the disentangled model by unsupervised learning without inductive biases both on the learning approaches and the data sets, or in other words, unsupervised DRL is impossible without inductive biases. 
Specifically, let us consider a generative paradigm as follows:

{\small
\begin{align}
p(\mathbf{x})=\int p(\mathbf{x}|\mathbf{z})p(\mathbf{z})d\mathbf{z}. 
\label{eq:identify}
\end{align}
}

\noindent The unsupervised DRL method has access to observations $\mathbf{x}$, i.e., $p(\mathbf{x})$, but it can not identify the true prior of latent variables $\mathbf{z}$ according to the marginal distribution $p(\mathbf{z})$. Theoretically, there is an infinite number of different distributions having the same $p(\mathbf{z})$~\cite{locatello2019challenging,khemakhem2020variational}. For example, if $p(\mathbf{z})$ is a multivariate Gaussian distribution, which will be invariant to rotation. Therefore, according to Eq.\eqref{eq:identify}, we will obtain infinite equivalent generative models which have the same $p(\mathbf{z})$. Let $\hat{\mathbf{z}}$ denote the latent variable of another generative model, so we have:

{\small
\begin{align}
p(\mathbf{x})=\int p(\mathbf{x}|\mathbf{z})p(\mathbf{z})d\mathbf{z}=\int p(\mathbf{x}|\hat{\mathbf{z}})p(\hat{\mathbf{z}})d\hat{\mathbf{z}} ,
\label{eq:identify:2}
\end{align}
}

\noindent where $p(\mathbf{z})$ equals to $p(\mathbf{\hat{z}})$. In summary, we can not ensure we actually obtain the disentangled model rather than other equivalent ones.

Therefore, we need extra inductive biases to identify the disentangled model, or we can leverage supervision to help find the target model~\cite{locatello2019challenging}. 
As for the aforementioned unsupervised methods, the reason why they can achieve DRL to some extent is that they also have inductive biases, e.g., the regularizers and their regularization strength. Locatello et al.~\cite{locatello2019challenging} also point out that the disentanglement scores of unsupervised DRL can be easily influenced by randomness and hyper-parameters.
As such, although designing appropriate inductive biases is important for unsupervised DRL, it might be more effective to use implicit and explicit supervision.

\subsection{Independence Assumption}
\label{ind vs. causal}

Intuitively, typical DRL models discussed so far hold the assumption that latent factors are statistically independent, so that they are supposed to be independently disentangled through independent or factorial regularization~\cite{higgins2016beta, kim2018disentangling} or various disentanglement losses~\cite{chen2023disenbooth,liu2021activity}.
However, in some cases, underlying generative factors are not independent and hold certain causal relations. In this section, we discuss causal DRL methods that can capture the underlying causal mechanism of the data generation process and potentially achieve more interpretable and robust representations via disentangling causal factors.


Based on the statement from Suter et al.~\cite{suter2019robustly}, Reddy et al.~\cite{reddy2021causally} propose two essential properties that a generative latent variable models (e.g., VAE) should fulfill to achieve causal disentanglement. 
Consider a latent variable model $M (e, g, p_X)$, where $e$ denotes an encoder, $g$ denotes a generator and $p_x$ denotes a data distribution. Let $G_i$ denote the $i$-th generative factor and $C$ be the confounders in the causal learning literature~\cite{greenland1999confounding}. The two properties with respect to encoder and generator are presented in the following:

\textbf{Property 1}. Encoder $e$ can learn the mapping from $G_i$ to unique $Z_I$, where $I$ is a set of indices and $Z_I$ is a set of latent dimensions indexed by $I$ . The unique $Z_I$ means that $Z_{I} \cap Z_{J}=\emptyset, \ \forall \ I \neq J,|I|,|J| \geq 0$. In this case, we assert that $Z$ is unconfounded with respect to $C$, i.e., there is no spurious correlation between $Z_I \ \text{and} \  Z_J, \ \forall \ I \neq J $.
 
\textbf{Property 2}. For a generative process by $g$ , only $Z_I$ can influence the aspects of generated output controlled by $G_i$, while the others, denoted as $Z_{I^-}$, can not.
 
\begin{figure}[htp]
    \centering
    \includegraphics[width=0.90\linewidth]{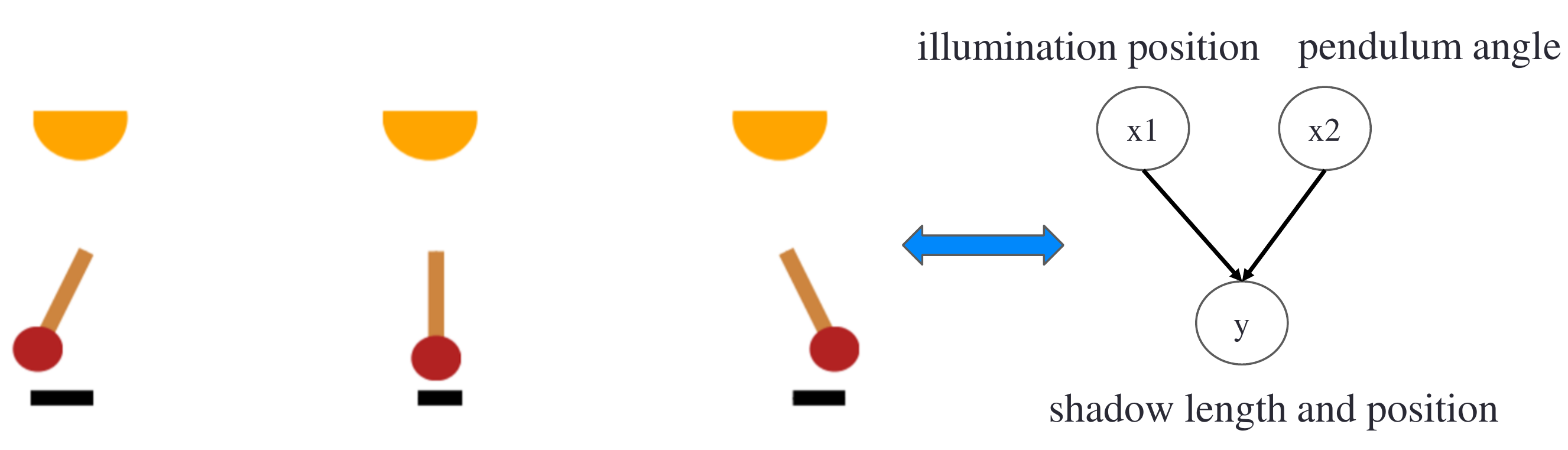}
    \caption{The position of the illumination source and the angle of the pendulum are causes of the position and the length of the shadow.}
    \label{fig:causal_vae}
\end{figure}

Since Locatello et al.~\cite{locatello2019challenging} challenge the common assumption in the vanilla VAE based DRL approaches that latent variables need to be independent, some following works also attempt to discard the independence assumptions. 
Yang et al.~\cite{Yang_2021_CVPR} propose CausalVAE which first introduces structural causal model (SCM) as prior. CausalVAE considers the relationships between the factors of variation in the data from the perspective of causality, 
describing these relationships with SCM, as is illustrated in Figure.~\ref{fig:causal_vae}. CausalVAE employs an encoder to map the input $\mathbf{x}$ and supervision signal $\mathbf{u}$ associated with the true causal concepts to an independent exogenous variable $\boldsymbol{\epsilon}$ whose prior distribution follows a standard Multivariate Gaussian $\mathcal{N}(\mathbf{0}, \mathbf{I})$. This encoding process is illustrated in Eq.~\eqref{eq:causalVAE3},

{\small
\begin{align}
\boldsymbol{\epsilon}=h(\mathbf{x}, \mathbf{u})+\zeta ,
\label{eq:causalVAE3}
\end{align}
}

\noindent where $h$ is the encoder and $\zeta$ is a noise. Then a \textit{Causal Layer} is designed to transforms $\boldsymbol{\epsilon}$ to causal representation $\mathbf{z}$ through the linear structural equation in Eq.\eqref{eq:causalVAE1},

{\small
\begin{align}
\mathbf{z}=\mathbf{A}^{\top} \mathbf{z}+\boldsymbol{\epsilon}=\left(\mathbf{I}-\mathbf{A}^{\top}\right)^{-1} \boldsymbol{\epsilon} ,
\label{eq:causalVAE1}
\end{align}
}

\noindent where $\mathbf{A}$ is the learnable adjacency matrix of the causal directed acyclic graph (DAG). Before being fed into the decoder, $\mathbf{z}$ is passed through a \textit{Mask Layer} to reconstruct itself, as is illustrated in Eq.\eqref{eq:causalVAE2}, for the $i$-th latent dimension of $\mathbf{z}$, $z_i$,

{\small
\begin{align}
z_{i}=g_{i}\left(\mathbf{A}_{i} \circ \mathbf{z} ; \eta_{i}\right)+\epsilon_{i},
\label{eq:causalVAE2}
\end{align}
}

\noindent where $\circ$ represents element-wise product and $g_i$ is a mild nonlinear function with the learnable parameter $\eta_i$. In this mask stage, causal intervention is conducted in the form of ``do operation'' by setting $z_i$ to a fixed value. After the \textit{Mask Layer}, $\mathbf{z}$ is passed through the decoder to reconstruct the observation $\mathbf{x}$, i.e., $\hat{\mathbf{x}}=\mathbf{d}(\mathbf{z})+\xi$, where $\xi$ is also a noise.

Bengio et al.~\cite{Bengio2020A} point out adaptation speed can evaluate how well a model fits the underlying causal structure from the view of causal inference, 
and exploit a meta-learning objective to learn disentangled and structured causal representations given unknown mixtures of causal variables.

Different from the supervised scheme of CausalVAE, Shen et al.~\cite{shen2020disentangled} propose a weakly supervised framework named DEAR, which also introduces SCM as prior. 
First, the causal representation $\mathbf{z}$ is obtained by an encoder E (or obtained by sampling from prior $p_z$), taking sample $\mathbf{x}$ as input, i.e., $\mathbf{z}=E(\mathbf{x})$.
Second, the exogenous variable $\boldsymbol{\epsilon}$ is computed based on the general non-linear SCM proposed by Yu et al.~\cite{yu2019dag} in which the previously calculated $\mathbf{z}$ is employed to define $F_{\beta}(\boldsymbol{\epsilon})$, as is shown in Eq.\eqref{eq:dear1},

{\small
\begin{align}
\Big[ \mathbf{z}=f_1\big((\mathbf{I}-\mathbf{A}^{\top})^{-1} f_2(\boldsymbol{\epsilon})\big) \Big]:= F_{\beta}(\boldsymbol{\epsilon}),
\label{eq:dear1}
\end{align}
}

\noindent where $f_1$ and $f_2$ are element-wise transformations, which are usually non-linear. $A$ is the same learnable adjacency matrix in Eq.\eqref{eq:causalVAE1} and Eq.\eqref{eq:causalVAE2}. $\beta$ denotes the parameters of $f_1$, $f_2$ and $A$. When $f_1$ is invertible, Eq.\eqref{eq:dear1} will be equivalent to Eq.\eqref{eq:dear2} in the following:

{\small
\begin{align}
f_1^{-1}(\mathbf{z})=\mathbf{A}^{\top} f_1^{-1}(\mathbf{z})+f_2(\boldsymbol{\epsilon}).
\label{eq:dear2}
\end{align}
}

\noindent Third, we can carry out "do operation" on $\mathbf{z}$ by setting $z_i$ to a fixed value and then reconstruct $\mathbf{z}$ using ancestral sampling by performing Eq.\eqref{eq:dear2} iteratively. 
Finally, $\mathbf{z}$ is passed through a decoder for reconstruction. To guarantee disentanglement, a weakly supervised loss $L=\mathbb{E}_{\mathbf{x},\mathbf{y}}\left[L_{s}(E;\mathbf{x},\mathbf{y})\right]$ is applied, only needing a small piece of labeled data, with $L_{s}=\sum_{i=1}^{m} \operatorname{CrossEntropy}\big(\bar{E}(x_{i}),y_{i}\big)$ when label $y_i$ is binarized or $L_{s}=\sum_{i=1}^{m}\big(\bar{E}(x_{i})-y_{i}\big)^{2}$ when $y_i$ is continuous. Note that $\bar{E}$ is the deterministic part of $E(\mathbf{x})$. When using the VAE structure, $\bar{E}(\mathbf{x})=m(\mathbf{x})$ is derived with $E(\mathbf{x})\sim \mathcal{N}\big(m(\mathbf{x})$, $\Sigma(\mathbf{x})\big)$, where $m(\mathbf{x})$ and $\Sigma(\mathbf{x})$ are the mean and variance output by the encoder, respectively.



\spara{Discussion}
Causal models in disentangled representation learning typically involve two main components:

\begin{itemize}
\item Structural Causal Models (SCMs): SCMs provide a way to represent causal relationships between variables using a directed acyclic graph (DAG). In an SCM, each node in the graph represents a variable, and the directed edges indicate causal dependencies. The variables can be observed or latent, and the model specifies how the variables interact with each other.

\item Interventional Inference: Causal models enable intervention, which means we can manipulate or intervene on specific variables to observe the effects on other variables. Interventions involve changing the value of a specific variable in the model and observing the resulting changes in the other variables. This helps in understanding the causal relationships and in encoding causal mechanisms into disentangled representations.
\end{itemize}

By incorporating causal models into disentangled representation learning, we can explicitly identify and disentangle the causal factors that influence the observed data. In our opinion, compared to independent DRL, causal DRL methods are better suited for scenarios characterized by the presence of multiple generating factors and potential causal relationships among these factors.  
Note that learning causal models and disentangled representations is a challenging task. In practice, it may be difficult to accurately specify the underlying causal structure and capture all the causal relationships in the data. Various techniques and algorithms, such as causal inference, causal discovery, and structural equation modeling, are employed in this area to address these challenges.

\subsection{The Interrelations of DRL}

In this section, we discuss the interrelations of DRL with other learning paradigms which have close relations with the idea of learning disentangled representations, i.e., Capsule Networks and Object-centric Learning. These two paradigms can be regarded as particular instances of DRL.

\subsubsection{The Interrelations with Capsule Nets}

Capsule networks~\cite{ribeiro2022learning,sabour2017dynamic,patrick2022capsule,mazzia2021efficient} introduced by Hinton et al.~\cite{sabour2017dynamic} are an alternative to traditional convolutional neural networks (CNNs), which aim to tackle the limitations of CNNs, e.g., the information loss brought by pooling, and the lack of ability to cope with part-object spatial relationships. Compared to scalar neural units, capsule networks organize the neurons into capsules, each of which is a group of neurons that work together to represent a specific feature, such as pose, color and texture.
The capsules can encode not only the presence of an object but also its properties and relationships with other objects. Capsule networks explicitly model the part-whole hierarchical relations through capsules in multiple levels. The lower capsules capture the information at the lower abstraction level, and the higher capsules capture the higher ones. Routing strategies are used to transmit information from the lower capsules to the higher capsules.

The concepts of capsule networks inherently coincide with those of DRL, as they tend to represent various factors of variation as separate capsules and decompose the features of objects into their composing parts. However, it can not be ensured that the representations of each capsule are indeed disentangled~\cite{hu2023beta}. Therefore it is still an open problem to obtain disentangled capsules. Hu et al.~\cite{hu2023beta} propose $beta$-CapsNet to learn disentangled capsules by adding information bottleneck constraints. The results show that $beta$-CapsNet has a better ability to learn disentangled representations than $beta$-VAE and the original CapsNet. In our humble opinions, capsule networks inherently have the potential to achieve DRL for their hierarchical part-whole architecture, providing a suitable framework for learning disentangled representations. Furthermore, it needs extra regularizers or explicit supervisions to ensure the disentanglement of capsules.
The combination of capsule networks and DRL has the potential to obtain more interpretable and robust representations of data, which is no doubt a promising research direction.

\subsubsection{The Interrelations with Object-centric Learning}

Conventional deep learning methods usually treat scenes as a whole, without explicitly considering individual objects and the compositional architectures.
In contrast, object-centric learning~\cite{locatello2020object,seitzer2022bridging,singh2022simple,elsayed2022savi++} aims to explicitly model and understand different objects individually, and capture the underlying structure of the scene and relationships of objects. For example, object-centric learning can output the masks of different objects and their object-centric representations that can be used in downstream tasks.
Object-centric learning highlights the significance of identifying and reasoning about objects as distinct entities, which can benefit a variety of downstream tasks such as controllable image generation~\cite{sylvain2021object}, segmentation~\cite{cheng2021boundary}, and visual question answering~\cite{wu2022slotformer}. Moreover, there are also a series of works that achieve more fine-grained object-centric learning, i.e., further disentangling the representations of each object. For example, Ferraro et al.~\cite{ferraro2022disentangling} disentangle shape and pose for each object. Li~\cite{li2020learning} disentangle several factors for each object such as rotation and color. In summary, object-centric learning can be regarded as a specific instance of DRL which focuses on the disentanglement of individual objects and their properties.

\subsection{Discussions on Connections over Taxonomy}

\red{We present the taxonomy of disentangled representation learning from four aspects: i) model type, ii) representation structure, iii) supervision signal, and iv) independence assumption, together with a lot of related works for each aspect. Although these works may focus on different aspects of DRL, we argue that these four aspects are interrelated, with the common assumption that there exist explainable factors hidden in the data and the final goal is to discover these factors. 
For instance, the independence assumption is able to influence the model design, where if the factors have causal relationships, we may choose a causal-VAE model instead of vanilla-VAE. Similarly, the representation structure also has an impact on the model type, where the dimension-wise representation structure is more suitable for VAE-based model. Future investigations may simultaneously consider these four aspects and conduct joint optimization over them, to achieve disentanglement on various target tasks. }

\section{Metrics}
\label{sec:metrics} 


Many works~\cite{higgins2016beta,kim2018disentangling,chen2019isolating,chen2016infogan} qualitatively evaluate the performance of disentanglement by inspecting the change in reconstructions when traversing one variable in the latent space. Qualitative observation is straightforward, but not precise or mathematically rigorous. In order to promote the research of learning disentangled representations, it is important to design reliable metrics which can quantitatively measure disentanglement. 
We review and divide a series of quantitative metrics into two categories: supervised metrics and unsupervised metrics. As for a deeper understanding, discussion and taxonomy for metrics, we refer interested readers to Zaidi et al.'s work~\cite{zaidi2020measuring}.

\subsection{Supervised Metrics}

Supervised metrics assume that we have access to the ground truth generative factors.


\textbf{Z-diff.} Higgins et al.~\cite{higgins2016beta} propose a supervised disentanglement metric based on a low capacity linear classifier network to measure both the independence and explainability. They conduct inference on a number of image pairs that are generated by fixing the value of one data generative factor while randomly sampling all others. Taking a batch of image pairs as input, the classifier is expected to identity which factor is fixed and report the accuracy value as the disentanglement metric score.

\textbf{Z-min Variance.} Kim et al.~\cite{kim2018disentangling} point out that the aforementioned method using linear network has several weaknesses, such as being sensitive to hyperparameters of the linear classifier optimization. Most importantly, the metric has a failure mode: giving 100\% accuracy even when only $K-1$ factors out of K have been disentangled. The authors propose a metric based on a majority-vote classifier with no optimization hyperparameters. They also generate a number of images with one factor $k$ fixed and all others varying randomly. After obtaining representations with normalization, they take the index of the dimension with the lowest empirical variance and the label $k$ as the input \& output for the majority-vote classifier. The accuracy of the classifier is regarded as the disentanglement metric score.

\textbf{Z-max Variance.} Kim et al.~\cite{kim2019relevance} propose a metric which is almost the same as Z-min Variance. The main difference lays that they generate samples with one factor $k$ varying and all others fixed. Consequently, they choose the index of the dimension with the highest empirical variance as the input of majority-vote classifier. They claim that this metric shows better consistency with qualitative assessment than Z-min Variance.

\textbf{Mutual Information Gap (MIG).} Chen et al.~\cite{chen2019isolating} propose a classifier-free information-theoretic metric named MIG. The key insight of MIG is to evaluate the empirical mutual information between a latent variable $z_j$ and a ground truth factor $k$. For each factor $k$, MIG computes the gap between the top two latent variables with the highest mutual information. The average gap over all factors is used as the disentanglement metric score. Higher MIG score means better disentanglement performance because it indicates that each generative factor is principally captured by only one latent dimension.

\textbf{SAP Score.}  Kumar et al.~\cite{kumar2018variational} propose a metric referred as Separated Attribute Predictability (SAP) score. They construct a score matrix $S \in {R}^{d \times k}$ and the $(i,j)-$th element represents the linear regression or classification score of predicting $j-$th factor using only $i-$th latent variable distribution. Then for each column of the score matrix, they compute the difference between the top two elements and take the average of these differences as the SAP score. Higher SAP score means better disentanglement performance because it also indicates that each generative factor is principally corresponding to only one latent dimension, just like MIG.

\textbf{DCI.} Eastwood et al.~\cite{eastwood2018framework} design a framework which evaluates disentangled models from three aspects, i.e., disentanglement (D), completeness (C) and informativeness (I). Specifically, disentanglement denotes the degree of capturing at most one generative factor for each latent variable. Completeness denotes the degree to which each generative factor is captured by only one latent variable. Informativeness denotes the amount of information that latent variables captures about the generative factors. It is worth noting that the disentanglement and the  completeness together quantify the deviation between bijection and the actual mapping.

\textbf{Modularity and Explicitness.} Ridgeway et al.~\cite{ridgeway2018learning} evaluate disentanglement from two aspects, i.e., modularity and explicitness. They claim a latent dimension is ideally modular only when it has high mutual information with only one factor and zero with all others. They obtain the modularity score by computing the deviation between the empirical case and the desired case. Explicitness focuses on the coverage of latent representation with respect to generative factors. Assuming factors have discrete values, they fit a one-versus-rest logistic-regression factor classifier and record the ROC area-under-the-curve (AUC). They then take the mean of AUC values over all classes for all factors as the final explicitness score.

\textbf {UNIBOUND.} Tokui et al.~\cite{tokui2022disentanglement} propose UNIBOUND to evaluate disentanglement by lower bounding the unique information in the term of Partial Information Decomposition (PID). PID decomposes the information between a latent variable $z_l$ and a generative factor $y_k$ into three parts: redundant information, unique information and complementary information. Let $\mathcal{U}(y_{k} ; z_{\ell} \backslash \mathbf{z}_{\backslash \ell})$ denote the unique information which is held by $z_{\ell}$ and not held by remaining variables, they then lower bound this unique information term. Similar to MIG, for each generative factor, the difference between the top two latent variables with largest lower bound value is computed. The average value over all factors is taken as the final score.

\textbf {UC and GC.} Reddy~\cite{reddy2021causally} propose Unconfoundedness (UC) Metric and Counterfactual Generativeness (CG) Metric from the causal perspective. As mentioned in Section~\ref{sec:definition}, they leverage an SCM to describe the data generation process. UC metric evaluates the degree how the mapping from $G_i$ to $Z_I$ is unique and unconfounded with respect to a set of confounders $C$. UC is defined as $U C:=1-\mathbb{E}_{x \sim p_{X}}\left[\frac{1}{S} \sum_{I, J} \frac{\left|\mathbf{Z}_{I}^{x} \cap \mathbf{Z}_{J}^{x}\right|}{\left|\mathbf{Z}_{I}^{x} \cup \mathbf{Z}_{J}^{x}\right|}\right]$. CG evaluates whether or not any causal intervention on $Z_{\\I}$ influence the generated aspects about $G_i$. This means only the intervention on $Z_I$ can influence $G_i$ for the generation process. CG is defined as $CG=\mathbb{E}_{I}[|A C E_{\mathbf{Z}_{I}^{X}}^{X_{I}^{c f}}-A C E_{\mathbf{Z}_{\backslash I}^{X}}^{X_{\backslash I}^{c f}}|]$, where $A C E_{d o(Z=\alpha)}^{X}=\mathbb{E}[X | d o(Z= \alpha)]-\mathbb{E} [X | d o (Z=\alpha^{*})]$.


\subsection{Unsupervised Metrics}

When we do not have access to the ground truth factors, unsupervised metrics then become important and useful. 


\textbf{ISI.} Do et al.~\cite{do2019theory} suggest three important properties of disentanglement from the perspective of mutual information, i.e., informativeness (I), separability (S) and interpretability (I). Furthermore, they propose a series of metrics to conduct the evaluation based on the three aspects respectively. Specifically, informativeness denotes the mutual information between original data $x$ and latent variable $z_i$, formulated as $I(x,z_i)$  Separability means that any two latent variables $z_i,z_j$ do not share common information about the data $x$, which denotes the ability to separate two latent variables with respect to the data $x$, formulated as $I(x,z_i,z_j)$. Explainability means a one-one mapping (or bijection) between latent variables $z_i$ and the data generative factors $y_k$, formulated as $I(z_{i},y_{k})=H(z_{i})=H(y_{k})$. They further propose specific methods of estimating these mutual information terms, which are applicable to both supervised and unsupervised scenarios.

\section{DRL Applications}
\label{sec:applications}

In this section, we discuss the broad applications of DRL for various downstream tasks. Firstly, we will discuss DRL applications of different modalities, e.g., image, video, natural language, and multi-modality, which almost cover all the application fields of deep learning. Besides, we introduce DRL applications in recommendation and graph learning. Table~\ref{summary of application} is a brief summary.
Finally, we demonstrate the effectiveness of DRL in few-shot and out-of-distribution (OOD) problems.

\subsection{Image}
Images, as one of the most widely investigated visual data types, can benefit a lot from DRL in terms of generation, translation, explanation, etc.

\subsubsection{Generation}

By taking advantages of DRL, independent factors in generation objectives can be learned and aligned with latent representation through disentanglement, hence capable of controlling the generation process.

On the one hand, the original VAE~\cite{kingma2013auto} model learns well-disentangled representations on image generation and reconstruction tasks. 
Later approaches have achieved more prominent results on image manipulation and intervene through improvement in disentanglement and reconstruction. 
Representative models such as ${\beta}$-VAE~\cite{higgins2016beta,burgess2018understanding} and FactorVAE~\cite{kim2018disentangling} can better disentangle independent factors of variation, enabling applicable manipulations of latent variables in the image generation process.
JointVAE~\cite{dupont2018learning} pays attention to joint continuous and discrete features, which acquires more generalized representations compared with previous methods, 
thus broadening the scope of image generation to a wider range of fields.
CausalVAE~\cite{Yang_2021_CVPR} introduces causal structure into disentanglement with weak supervision, supporting the generation of images with causal semantics and creation of counterfactual results.


On the other hand, GAN-based disentangled models have also been applied in image generation tasks, benefiting in the high fidelity of GAN. 
InfoGAN~\cite{chen2016infogan}, as a typical GAN based model, disentangles latent representation in an unsupervised manner to learn explainable representations and generates images under manipulation, 
while lacking of stability and sample diversity~\cite{higgins2016beta,kim2018disentangling}.
Larsen et al.~\cite{larsen2016autoencoding} combine VAE and GAN as an unsupervised generative model by i) merging the decoder and the generator into one, ii) using feature-wise similarity measures instead of element-wise errors, which learns high-level visual attributes for image generation and reconstruction in high fidelity, iii) suggesting that unsupervised training produces certain disentangled image representations.
Zhu et al.~\cite{zhu2018visual} utilize GAN architecture to disentangle 3D representations including shape, viewpoint, and texture, to synthesize natural images of objects.
Wu et al.~\cite{wu2021stylespace} analyze disentanglement generation operation in StyleGAN~\cite{Karras_2019_CVPR}, especially in \textit{StyleSpace}, to manipulate semantically meaningful attributes in generation.
Zeng et al.~\cite{zeng2020realistic} propose a hybrid model DAE-GAN, which utilizes a deforming autoencoder and conditional generator to disentangle identity and pose representations from video frames, 
generating realistic face images of particular poses in a self-supervised manner without manual annotations.

Other works based on information theory also make considerable contributions for long. 
For example, Gao et al. propose InfoSwap~\cite{Gao_2021_CVPR}, which disentangles identity-relevant and identity-irrelevant information through optimizing information bottleneck to 
generate more identity-discriminative swapped faces.

\blue{
\subsubsection{Domain Adaption}
In addition to generation, domain adaption is also a hot topic in image processing and understanding. 
Disentangled factors contribute to coherent and robust performance for cross-domain scenarios, 
ultimately enhancing and expanding the controllability and applicability of domain adaption. DRL is usually used to disentangle the domain-invariant and the domain-specific factors to help domain adaption tasks.

Gonzalez et al.~\cite{gonzalez2018image} present cross-domain disentanglement, disentangling the internal representations into shared and exclusive parts through bidirectional image translation based on GAN and cross-domain autoencoders with only paired images as input. 
This design achieves satisfactory performance on various tasks such as diverse sample generation, cross-domain retrieval, domain-specific image transfer and interpolation.
Lee et al.~\cite{Lee_2018_ECCV} disentangle latent representations into domain-invariant content space and domain-specific attribute space by introducing a content discriminator and cross-cycle consistency loss on GAN-based framework, achieving diverse multimodal translation without using pre-aligned image pairs for training.
Later, DRANet~\cite{Lee_2021_CVPR} is proposed to disentangle content and style factors, and synthesize images by transferring visual attributes for unsupervised multi-directional domain adaption.
Liu et al.~\cite{Liu_2021_CVPR} point out the lack of graduality for existing image translation models in semantic interpolations both within domains and across domains. 
As such, they propose a new training protocol, which learns a smooth and disentangled latent style space to perform gradual changes and better preserve the content of the source image.
}


\subsubsection{Others}
The idea of DRL has also been employed in other image-related fields and tasks.
Sanchez et al.~\cite{sanchez2020learning} disentangle shared and exclusive representations in paired images through optimizing mutual information, which is well applied to image classification and image retrieval tasks without relying on image reconstruction or image generation. 
Hamaguchi et al.~\cite{Hamaguchi_2019_CVPR} propose a VAE-based network to disentangle variant and invariant factors for rare event detection on imbalanced datasets, requiring only pairs of observations.
Gidaris et al.~\cite{gidaris2018unsupervised} propose a self-supervised semantic feature learning method through predicting rotated images with ConvNet model to achieve comparable performances with supervised methods.
Inspired by Gidaris et al.'s work, Feng et al.~\cite{Feng_2019_CVPR} later disentangle feature representations relevant to semantic rotation and irrelevant ones through joint training on image rotating prediction and instance discrimination, which benefits in the generalization ability in image classification, retrieval, segmentation and other tasks.
Ghandeharioun et al.~\cite{ghandeharioun2021dissect} propose DISSECT, which enforces the disentanglement of latent concepts by encouraging the distinctness across different concepts and the proximity within a same concept. 
They achieve multiple counterfactual image explanations which can intervene the output of model by changing disentangled concepts.

To summarize, representations learned through various image representation models can always be structured based on DRL strategy to separate variant events from the inherent attributes. 
Therefore, as an appropriate learning strategy for image-related tasks, DRL particularly contributes to improvement in image generation and translation, facilitating more comprehensive and diverse implementations for various image applications.

\subsection{Video}
Besides static images, DRL also promotes dynamic videos analysis, including video prediction, video retrieval and motion retargeting etc.

\subsubsection{Video Prediction}

Video prediction is a challenging yet interesting task of predicting future frames given a set of contextual frames.
Denton et al. propose DRNET~\cite{denton2017unsupervised}, an autoencoder-based model factorizing each frame into an invariant part and a varying component, which is able to coherently generate future frames in videos.
One of the major challenges for video prediction lays in the high dimensional representation space of visual data. 
To tackle this problem, Sreekar et al. propose mutual information predictive auto-encoder (MIPAE)~\cite{sreekar2021mutual}, separating latent representations into time-invariant (content) and time-varying (pose) part, 
which avoids directly predictions of high dimensional video frames. They use a mutual information loss and a similarity loss to enforce disentanglement, as well as employ LSTM to predict low dimensional pose representations.
Latent representations of content and the predicted representations of pose are then decoded to generate future frames.
Hsieh et al. later propose DDPAE~\cite{hsieh2018learning}, a framework which also disentangles the content representations and the low-dimensional pose representations. 
They utilize a pose prediction neural network to predict future pose representations based on the existing pose representations. Based on an inverse spatial transformer parameterized by the predicted pose representations, 
the invariant content representations can also be used to predict future frames.



\subsubsection{Others}

To deal with the large mode variations in the real-world applications, Kim et al.~\cite{kim2020robust} propose a DRL framework for robust facial authentication, which disentangles identity and mode (e.g., illumination, pose) features. They first use two encoders to encode identity and mode, respectively. To ensure disentanglement, they design an exclusion-based strategy which encourages the two encoders to remove the characteristics of the peer from their own representations. Moreover, they design a reconstruction-based strategy to reinforce the disentanglement, which uses a decoder to reconstruct the original features by exchanging identity features for an image-pair before re-disentangling the identity and the mode features.


Motion retargeting aims at transferring human motion from a source video to a target video. Ma et al.~\cite{ma2022human} point out that previous motion retargeting methods neglect the subject-dependent motion features in the transferring process, which leads to unnatural synthesis. To tackle this problem, they propose to disentangle subject-dependent motion features and subject-independent motion features, generating target motion features through combination of the source subject-independent features and the target subject-dependent features. To achieve this purpose, they design triplet loss for both subject-dependent and subject-independent features to ensure the disentanglement.


\subsection{Natural Language Processing}
DRL has also been explored in natural language processing (NLP) tasks, such as text generation, style transfer, semantic understanding etc.

\subsubsection{Text Representation}
The initial DRL applications in NLP aims at learning disentangled text representations w.r.t. various criteria, primarily by encoding different aspects of representations into distinct spaces. 
He et al.~\cite{he2017unsupervised} apply attention mechanism to an unsupervised neural word embedding model so as to discover meaningful and semantically coherent aspects with strong identification, which improves disentanglement among diverse aspects compared with previous approaches.
Bao et al.~\cite{bao2019generating} generate sentences from disentangled syntactic and semantic spaces through modeling syntactic information in the latent space of VAE and regularizing syntactic and semantic spaces via an adversarial reconstruction loss.
Cheng et al.~\cite{cheng2020improving} propose a disentangled learning framework with partial supervision for NLP, to disentangle the information between style and content of a given text by optimizing the upper bound of mutual information. With the semantic information being preserved, this framework performs well on conditional text generation and text-style transfer.
Wu et al.~\cite{wu2020improving} propose a disentangled learning method that optimizes the robustness and generalization ability of NLP models.
Colombo et al.~\cite{colombo2021novel} propose to learn disentangled representation for text data
by minimizing the mutual information between the latent representations of the sentence contents and the attributes. They design a novel variational upper bound based on the Kullback-Leibler and the Renyi divergences to estimate the mutual information.

\subsubsection{Style Transfer}

Several works are motivated by employing DRL to disentangle style information from text representations in the practice of text style transfer tasks.
Hu et al.~\cite{hu2017toward} combine VAE with an attribute discriminator to disentangle content and attributes of the given textual data, for generating texts with desired attributes of sentiment and tenses.
John et al.~\cite{john2019disentangled} incorporate auxiliary multi-task and adversarial objectives based on VAE to disentangle the latent representations of sentence, achieving high performance in non-parallel text style transfer.

\subsubsection{Others}
There also exist specific tasks in NLP community where DRL serves as an effective approach.
Zou et al.~\cite{zou2022divide} propose to address the fundamental task, i.e., text semantic matching, by disentangling factual keywords from abstract to learn the fundamental way of content matching under different levels of granularity.
Dougrez-Lewis et al.~\cite{dougrez2021learning} disentangle the latent topics of social media messages through an adversarial learning setting, to achieve rumour veracity classification.
Zhu et al.~\cite{zhu2021neural} disentangle the content and style in latent space by diluting sentence-level information in style representations to generate stylistic conversational responses.
Other works~\cite{zhang2021disentangling, zeng2022task} also propose to exploit the large pretrained language models (PLM) using DRL.
Zhang et al.~\cite{zhang2021disentangling} try to uncover disentangled representations from pretrained models such as BERT~\cite{devlin2019bert} by identifying existing subnetworks within them, aiming to extract representations that can factorize into distinct, complementary properties of input.
Zeng et al.~\cite{zeng2022task} propose task-guided disentangled tuning for PLMs, which enhances the generalization of representations by disentangling task-relevant signals from the entangled representations.

\subsection{Multimodal Application}

With the fast development of multimodal data, there have also been an increasing number of research interests on DRL for multimodal tasks, where DRL is primarily conductive to the separation, alignment and generalization of representations of different modalities.

Early works~\cite{hsu2018disentangling, tsai2018learning} study the typical modal-level disentanglement through encouraging independence between modality-specific and multimodal factors.
Shi et al.~\cite{shi2019variational} posit four criteria for multimodal generative models and propose a multimodal VAE using a mixture-of-experts layer, achieving disentanglement among modalities.
Zhang et al.~\cite{zhang2022learning} propose a disentangled sentiment representation adversarial network (DiSRAN) to reduce the domain shift of expressive styles for cross-domain sentiment analysis.
Recent works~\cite{Alaniz_2022_CVPR, Xu_2022_CVPR, yu2022towards, Materzynska_2022_CVPR, zou2022utilizing} tend to focus on disentangling the rich information among multi modalities and leveraging that to perform various downstream tasks.
Alaniz et al.~\cite{Alaniz_2022_CVPR} propose to use the semantic structure of the text to disentangle the visual data, in order to learn an unified representation between the text and image.
The PPE framework~\cite{Xu_2022_CVPR} realizes disentangled text-driven image manipulation through exploiting the power of the pretrained vision-language model CLIP~\cite{radford2021learning}. Similarly, Yu et al.~\cite{yu2022towards} achieve counterfactual image manipulation via disentangling and leveraging the semantic in text embedding of CLIP.
Materzynska et al.~\cite{Materzynska_2022_CVPR} disentangle the spelling capabilities from the visual concept processing of CLIP.

\subsection{Recommendation}
Application of DRL in recommendation tasks has also drawn researchers' attention substantially. Latent factors behind user's behaviors can be complicated and entangled in recommender systems.  Disentangled factors bring new perspectives, reduce the complexity and  improve the efficiency and explainability of recommendation.

DRL in recommendation mostly aims at capturing user's interests of different aspects. Early works~\cite{ma2019learning, wang2020disentangled, zhang2020content,news} focus on learning disentangled representations for collaborative filtering. Specifically,
Ma et al.~\cite{ma2019learning} propose MacridVAE to learn the user's macro and micro preference on items, which can be used for controllable recommendation. Wang et al.~\cite{wang2020disentangled,news} decomposes the user-item bipartite graph into several disentangled subgraphs, indicating different kinds of user-item relations. Zhang et al.~\cite{zhang2020content} propose to learn users' disentangled interests from both behavioral and content information. More recent works~\cite{ma2020disentangled, chen2021curriculum,zhang2023adaptive} also applied DRL in the sequential recommendations, where the user's future interest are matched with historical behaviors in the disentangled intention space. Additionally, some works~\cite{wang2022disentangled, wang2021multimodal} also utilize auxiliary information to help the disentangled recommendation. In particular, Wang et al.~\cite{wang2021multimodal} utilize both visual images and textual descriptions to extract the user interests, providing recommendation explainability from the visual and textual clues. 
Later they incorporate both visual and categorical information to provide disentangled visual semantics which further boost both recommendation explainability and accuracy~\cite{wang2022disentangled}. Wang et al.~\cite{wang2023curriculum} further learn co-disentangled representation across different envrionments for social recommendation.

\subsection{Graph Representation Learning}
Graph representation learning and reasoning methods are being significantly demanded due to increasing applications on various domains dealing with graph structured data, while real-world graph data always carry complex relationships and interdependency between objects~\cite{zhang2020deep,zhou2020graph}. Consequently, research efforts have been devoted to applying DRL to graphs, resulting in beneficial advances in graph analysis tasks.

Ma et al.~\cite{ma2019disentangled} point out the absence of attention for complex entanglement of latent factors contemporaneously and proposes DisenGCN, which learns disentangled node representations through \textit{neighborhood routing mechanism} iteratively segmenting the neighborhood according to the underlying factors.
Later, NED-VAE~\cite{guo2020interpretable} is proposed to be one unsupervised disentangled method that can disentangle node and edge features from attributed graphs.
FactorGCN~\cite{yang2020factorizable} is then proposed to decompose the input graph into several factor graphs for graph-level disentangled representations. After that, each of the factor graphs is separately fed to the GNN model and then aggregated together for disentangled graph representations.
Li et al.~\cite{li2021disentangled} propose to learn disentangled graph representations with self-supervision and for out-of-distribution generalization~\cite{li2024disentangled}. Given the input graph, the proposed method DGCL identifies the latent factors of the input graph and derives its factorized representations. Then it conducts factor-wise contrastive learning to encourage the factorized representations to independently reflect the expressive information from different latent factors. 
They later propose IDGCL~\cite{li2022disentangled} that is able to learn disentangled self-supervised graph representation via explicit enforcing independence between the latent representations to improve the quality of disentangled graph representations. 
Zhang et al.~\cite{zhang2024disentangled} further apply DRL to continual graph neural architecture search.


\subsection{Abstract Reasoning}
\red{
Abstract reasoning in neural networks is proposed by Barrett et al.~\cite{barrett2018measuring}. Inspired by the human IQ test Raven’s Progressive Matrices (RPMs), they explore ways to measure and induce abstract reasoning ability as well as the generalization performance in neural networks.
Particularly, Steenkiste et al.~\cite{van2019disentangled} investigate the utility of DRL for abstract reasoning tasks. They evaluate the usefulness of the representations learned by DRL models and train abstract reasoning models based on these disentangled representations. They observe that disentangled representations may lead to better downstream performance, e.g., learning faster with fewer samples in tasks similar to RPMs.
Locatello et al.~\cite{locatello2020weakly} learn disentangled representations from pairs of observations, and present adaptive group-based disentanglement methods without requiring annotations of the groups. They demonstrate that disentangled representations can be learned with only weak supervision, which exhibits capabilities beyond statistical correlations and shows effectiveness over abstract visual reasoning tasks.
Amizadeh et al.~\cite{amizadeh2020neuro} investigate the disentanglement between the reasoning and perception for visual question answering (VQA). They introduce a differentiable first-order logic formalism for VQA that explicitly decouples logical reasoning from visual perception in the process of question answering, 
which enables the independent evaluation of reasoning and perception.
}

\begin{table*}[htb]
\centering
\small
\caption{Representatives of disentangled representation learning applications}
\begin{tabular}{l|l|l|l}
\hline
\multicolumn{1}{c|}{\textbf{Papers}} & \multicolumn{1}{c|}{\textbf{Model}} & \multicolumn{1}{c|}{\textbf{Paradigm}} & \multicolumn{1}{c}{\textbf{Application}} \\ 
\hline
    \cite{kingma2013auto, chen2019isolating, kumar2018variational, higgins2016beta, burgess2018understanding, kim2018disentangling, kim2019relevance, dupont2018learning}    & VAE-based     & \multirow{2}{*}{Unsupervised} & \multirow{5}{*}{Image generation}                          \\ 
\cline{1-2} 
    \cite{chen2016infogan, larsen2016autoencoding, zhu2018visual, wu2021stylespace}                                                                                          & GAN-based     &                               &                                                            \\ 
\cline{1-3} 
    \cite{kulkarni2015deep,bouchacourt2018multi,Locatello2020Disentangling, trauble2021disentangled}                                                                         & VAE-based     & \multirow{3}{*}{Supervised}   &                                                            \\ 
\cline{1-2} 
    \cite{tran2017disentangled, xiao2017dna}                                                                                                                                 & GAN-based     &                               &                                                            \\ 
\cline{1-2}
    \cite{Yang_2021_CVPR,shen2020disentangled}                                                                                                                               & Causal-based  &                               &                                                            \\ 
\hline
    \cite{gonzalez2018image, Lee_2018_ECCV, Lee_2021_CVPR, Liu_2021_CVPR,zeng2020realistic}                                                                                  & GAN-based     & Unsupervised                  &  Image translation                                         \\ 
\hline 
    \cite{Hamaguchi_2019_CVPR}                                                                                                                                               & VAE-based     & Supervised                    &  \multirow{2}{*}{Image classification, segmentation, etc.} \\
\cline{1-3}
    \cite{Feng_2019_CVPR,sanchez2020learning}                                                                                                                                & Others        & Unsupervised                  &                                                            \\ 
\hline  
    \cite{denton2017unsupervised,hsieh2018learning}                                                                                                                          & VAE-based     & Unsupervised                  &  \multirow{2}{*}{Video}                                    \\
\cline{1-3}
    \cite{liu2021activity}                                                                                                                                                   & Others        & Supervised                    &                                                            \\ 
\hline
    \cite{he2017unsupervised}                                                                                                                                                & Others        & Unsupervised                  &  \multirow{2}{*}{Natural language processing}              \\ 
\cline{1-3}
    \cite{cheng2020improving, wu2020improving}                                                                                                                               & Others        & Supervised                    &                                                            \\ 
\hline
    \cite{hsu2018disentangling, shi2019variational, Alaniz_2022_CVPR}                                                                                                        & VAE-based     & Unsupervised                  &  \multirow{4}{*}{Multimodal Application}              \\ 
\cline{1-3}
    \cite{tsai2018learning, zhang2022learning}                                                                                                                               & VAE-based     & \multirow{3}{*}{Supervised}   &                                                            \\
\cline{1-2}
    \cite{yu2022towards}                                                                                                                                                     & GAN-based     &                               &                                                            \\ 
\cline{1-2}
    \cite{Xu_2022_CVPR, Materzynska_2022_CVPR, zou2022utilizing}                                                                                                             & Others        &                               &                                                            \\ 
\hline
    \cite{ma2019learning, wang2021multimodal, ma2020disentangled, chen2021curriculum, wang2022disentangled}                                                                  & VAE-based     & \multirow{2}{*}{Supervised}   &  \multirow{2}{*}{Recommendation}                           \\ 
\cline{1-2}
    \cite{zhang2020content, wang2020disentangled, news}                                                                                                                      & Others        &                               &                                                            \\ 
\hline
    \cite{guo2020interpretable, yang2020factorizable, li2021disentangled, li2022disentangled}                                                                     & VAE-based     & \multirow{2}{*}{Supervised}    &  \multirow{2}{*}{Graph}                                    \\ 
\cline{1-2}
    \cite{ma2019disentangled, wang2020disentangled, li2022ood}                                                                                                               & Others        &                               &                                                            \\ 
\hline
\end{tabular}
\label{summary of application}

\end{table*}

\blue{
\subsection{Few-shot Learning}

Few-shot leaning (FSL) methods are mainly categorized into three types: gradient-based methods, data augmentation-based methods, and metric learning-based methods. DRL also has a series of applications in few-shot leaning, given its capablility of obtaining disentangled representations and making the learned representations more robust and generalized.

\subsubsection{Data Augmentation}

An important way of handling few-shot learning is via data augmentation. 
To tackle this issue that the extracted intra-class variance features might simultaneously contain certain class-discriminative features and irrelevant information to the novel sample, Xu et al.~\cite{xu2021variational} propose a novel disentangled data augmentation framework which captures intra-class variation features $z_V$ and class-discriminative $z_I$ feature independently. They use a VAE to extract $z_V$ from input images and reconstruct with the feature $z=z_I+z_V$. The distribution of $z_V$ is shared across all categories whose prior is $N(0,I)$. $z_I$ is obtained by max-pooling and a classifier is trained to guarantee that $z_I$ can truly capture class-discriminative information. 
Finally, the process of data augmentation is achieving by sampling multiple $z_V$.

Lin et al.~\cite{lin2021joint} propose a data augmentation framework for few-shot image classification using the intra-class variation learned from base classes. They also disentangle images into class-specific (class-discriminative) features and appearance-specific (intra-class variation) features. They generate additional samples for novel classes by combining the class-specific feature of the seed sample with multiple appearance-specific features extracted from arbitrary base samples.

\subsubsection{Metric Learning}

Cheng et al.~\cite{cheng2021disentangled} propose a disentangled metric learning based method for few-shot image classification, namely DFR. They use two encoders $E_{cls}$ and $E_{var}$ to extract class-specific and class-irrelevant features (e.g., background and image style) respectively. The class-specific feature is also called as discriminative feature which is finally used for few-shot classification, while the class-irrelevant feature is superfluous for classification. To encourage disentanglement, they use a gradient reverse layer (GRL)~\cite{ganin2015unsupervised} and a class discriminator to minimize the class-specific information captured by $E_{var}$. Moreover, the reconstruction loss and classification loss can also promote disentanglement.

Tokmakov et al.~\cite{tokmakov2019learning} propose a disentangled representation framework for few-shot image classification. They collect category-level attribute annotations such as  object parts (e.g., wing color) and scene elements (e.g., grass). Subsequently, they disentangle image representations into parts corresponding to different attributes. They expect that the image representations are equal to the sum of all the attribute representations and thus use a certain distance loss function. An orthogonality loss function is also used to encourage disentanglement. They demonstrate that the disentangled representation can generalize better to novel classes with fewer examples.

Prabhudesai et al.~\cite{prabhudesai2020disentangling} propose a disentangled 3D prototypical network, D3DP-Nets, for few-shot concept learning. They first encode RGB-D images to 3D feature maps by training an encoder-decoder. They then disentangle the 3D feature map into 3D shape codes and 1D style codes through two encoders. Based on the shape codes and the style codes, the model can compute prototypes for each concept. Because of the disentanglement property, the classifier can only focus on essential features for each concept and thus has better generalization ability with fewer samples. 

\subsubsection{others}
Sun et al.~\cite{sun2017learning} propose a disentangled framework, SA-VAE, for stylized font generation in the few-shot setting. They disentangle character input into content and style features. They use a style inference network to encode style code and use a content recognition network to extract content information. In the training, they design an intercross pair-wise optimization algorithm to enforce that the style encoder can purely capture style information, or in other words, different characters with the same style can have the same style code. By inferring the style representation, the model has a powerful few-shot generalization ability for unseen styles.

Guo et al.~\cite{guo2021learning} propose a disentangled framework for few-shot 
visual question answering. The key insight of their work is that separating novel objects into parts that have been learned can benefit the learning of these novel objects. Therefore, they design an attribute network to disentangle the attributes of answers. In the few-shot stage, the network constrains the answer by minimizing the distance between the answer and the sum of its attributes. Although the training samples of novel answers are few, the attributes of the novel answers have been learned in the base stage. Thus, the framework has effective generalization ability in the few-shot setting.

\subsection{Out-of-distribution Problems}

DRL holds the promising ability to learn more robust representations that can be generalized to out-of-distribution (OOD) scenarios, as it can separate the essential features related to task objectives from irrelevant ones.

Li et al.~\cite{li2022ood} find that learning disentangled graph representation can improve the out-of-distribution (OOD) generalization ability of GNNs. The proposed OOD-GNN model encourages the graph representation disentanglement by eliminating the statistical dependence among all dimensions of the output representation through iteratively optimizing the sample graph weights and graph encoder. 
The experiments covering realistic and challenging cases of graph distribution shifts show that the disentangled OOD-GNN model achieves significant OOD generalization improvement against representative state-of-the-art GNN methods.

Dittadi et al.~\cite{dittadi2020transfer} propose a framework for evaluating the out-of-distribution generalization of disentangled representations in downstream tasks. They design factor regression tasks and construct a variety of OOD settings. The results suggest that disentangled representations can improve the OOD performance in some cases.

Yoo et al.~\cite{yoo2023disentangling} propose a disentangled framework to tackle the out-of-distribution generalization against degree-related distributional shifts of directed network embedding. They assign six different factors for each directed edge and learn corresponding disentangled embedding for each node. The experiments demonstrate the disentangled embeddings promote out-of-distribution generalization
against various degrees of shifts in degree distributions

Zhang et al.~\cite{zhang2022towards} propose a DRL framework for domain generalization.
They disentangle domain-invariant factors and domain-specific factors by separating them into different sub-spaces. They only use the domain-invariant features to conduct prediction. The experiments show the disentangled representations promote OOD generalization in various datasets.

Mu et al.~\cite{mu2021sdf} also propose a DRL framework for articulated shape representation. They achieve disentanglement via saparating shape representations and articulation representations. With these disentangled representations, they can generate shapes for unseen instances with unseen articulation angles, which demonstrates the generalization ability of these representations in OOD setting.
}

\section{DRL Design  for Different Tasks}
\label{sec:designs}


In this section, we discuss commonly adopted strategies for DRL in practical applications, providing inspirations on designing various DRL models for specific tasks.
We summarize two key aspects for designing a DRL model: i) designing an appropriate representation structure according to a specific task, and ii) designing corresponding loss functions which force the representation to be disentangled without losing task-specific information.

\subsection{Design of Representation Structure} 
\label{sec:architecture}

Given a specific task, we first need to work out the structure of the disentangled representations. To this end, we should consider the structure of the underlying generative factors. Specifically, on the one hand, we should consider whether the factors have a hierarchical or flat structure. If it is the former, we should adopt hierarchical DRL methods, otherwise use the flat DRL. On the other hand, we should consider the number and the granularity of the generative factors that we need to disentangle. As discussed in Sec~\ref{discuss: dim vs. vec}, dimension-wise DRL methods are suitable for multiple fine-grained factors in simple scenarios, while vector-wise DRL methods are suitable for several coarse-grained factors in more complex and real-world scenarios. In this section, we discuss the designs of the representation structure of DRL according to the taxonomy of ``dimension-wise or vector-wise methods".

Consider the two complementary structures: i) dimension-wise: use a whole vector representation $\mathbf{z}$, which is fine-grained. ii) vector-wise: use two or more independent vectors $\mathbf{z}_1,\mathbf{z}_2...$ to represent different parts of data features, which is coarse-grained. To guarantee the disentanglement property, method i) usually requires that $\mathbf{z}$ is dimension-wise independent, while method ii) usually requires that $\mathbf{z}_i$ is independent with $\mathbf{z}_j$ where $i \neq j$.

If we choose dimension-wise methods for our application, typical models that we can select are the various VAE-based and GAN-based methods which have been elaborated in Section~\ref{sec:methods}. 
In this case, we can use VAE or GAN as our backbone and design extra loss functions to adapt to specific tasks. 
We can also use other model architectures, for example, InfoSwap~\cite{Gao_2021_CVPR} which uses a multi-layer encoder to extract task-relevant features and compresses the features layer by layer based on information bottleneck to discard task-irrelevant features. 

As for vector-wise methods, there are usually two ways of obtaining multiple latent vectors: i) preset these vectors or ii) employ different encoders which take original representations as input to separate the original whole vector into several different vectors. 
For example, DR-GAN~\cite{tran2017disentangled} explicitly sets a latent representation to represent pose and uses an encoder to extract identity code from input images, then leverages a supervised loss function to guarantee that the pose code and the identity code can really capture the pose and the identity information correspondingly.
Liu et al.~\cite{liu2021activity} leverage two encoders, namely motion encoder and appearance encoder, to respectively extract the motion feature and the appearance feature by passing through the original representation.
Cheng et al.~\cite{cheng2021disentangled} utilize two encoders $E_{cls}$ and $E_{var}$ to extract class-specific and class-irrelevant features, respectively. 
DRNET~\cite{denton2017unsupervised} also uses two encoders to extract the pose feature and content feature, respectively. 
DRANet~\cite{Lee_2021_CVPR} employs only one encoder to extract the content feature and then obtains the style feature by subtracting the content feature from the original feature. 
Similar to DRANet, Wu et al.~\cite{wu2021vector} adopt one encoder to extract domain-invariant features from an image feature map, followed by obtaining domain-specific features through subtracting domain-invariant features.


We have to point out that no matter which model structure is chosen, appropriate loss functions must be designed to guarantee that the representation is disentangled without losing the information carried in the data.

\subsection{Design of Loss Function}
\label{sec:loss}


Here, we will discuss the design of loss functions which enforce disentanglement and informativeness according to different model types, i.e., generative model and discriminative model. 
Overall, we summarize loss functions as $\mathcal{L}= \lambda_{1} \mathcal{L}_{re}+ \lambda_{2}\mathcal{L}_{disen}+\lambda_{3}\mathcal{L}_{task}$, where $\mathcal{L}_{re}$ denotes reconstruction loss, $\mathcal{L}_{disen}$ denotes disentanglement loss, 
and $\mathcal{L}_{task}$ denotes specific task loss.

The reconstruction loss, which is always essential for generation tasks, ensures that the disentangled representation are semantically meaningful and can recover the original data.
The disentanglement loss enforces the disentanglement of the representation. Moreover, reconstruction loss can sometimes facilitate disentanglement, as it expects the model to correctly reconstruct the data by these disentangled features.
The task loss is directly related to the task objective. In the cases that we jointly optimize the task loss and the disentangled module, the task loss can usually provide guidance for disentanglement. In contrast, if we adopt a two-stage scheme, i.e., firstly training the disentangled module and then applying the disentangled features in downstream tasks, the task loss can't guide the disentangling process.

\subsubsection{Generative task}

\begin{table*}[htp]
\centering
\caption{The summary of loss functions of several generative and discriminative tasks.}
\label{tab: loss functions}
\small
\begin{tabular}{|m{3cm}<{\centering}|m{3cm}<{\centering}|m{4.5cm}<{\centering}|m{3.5cm}<{\centering}|} 
\hline
Methods   & reconstruction loss  & disentanglement loss    & task loss  \\ 
\hline
VAE-based Approaches & VAE loss  & extra regularizers added to ELBO   & -    \\ 
\hline
InfoGAN~\cite{chen2016infogan}   & GAN loss &  $\text { maximizing } \lambda I(c ; G(z, c))$  & -      \\ 
\hline
DRANet~\cite{Lee_2021_CVPR}   & L1 loss  & perceptual loss, consistency loss   &  adversarial loss  \\ 
\hline
DRNET~\cite{denton2017unsupervised}  & L2 loss   & similarity loss, adversarial loss & -   \\ 
\hline
InfoSwap~\cite{Gao_2021_CVPR}  & consistency loss & information-compression loss   &  adversarial loss    \\
\hline

MAP-IVR~\cite{liu2021activity}   & L2 loss        &  $\mathcal{L}_{\text {orth }}=\cos \left(m^{v}, a^{v}\right)$,\ $\mathcal{L}_{class}$  & - \\
\hline

Hamaguchi et al.~\cite{Hamaguchi_2019_CVPR}  & VAE loss   & similarity loss, activation loss   &  classification loss   \\
\hline


Cheng et al.~\cite{cheng2021disentangled} & L1 loss   &  discriminative Loss   &  classification loss  \\
\hline 
Wu et al.~\cite{wu2021vector} & -   &  orthogonality loss, adversarial loss   & detection loss  \\
\hline

\end{tabular}
\end{table*}



Various VAE-based models mentioned in Section~\ref{sec:methods} all have explicit reconstruction loss included in ELBO and also utilize extra regularizers as disentanglement loss. 
As for GAN-based methods, the adversarial loss can be regarded as reconstruction loss as well, and the disentanglement loss can be mutual information constraints such as those adopted in InfoGAN~\cite{chen2016infogan} and IB-GAN~\cite{jeon2021ib}. 

DRANet~\cite{Lee_2021_CVPR} adopts a $L_1$ loss as the reconstruction loss. It uses an adversarial loss as the task loss to ensure the task objective (i.e., image cross-domain adaption) is satisfied. This task loss is computed according to the generated images based on disentangled features, so it also encourages the model to obtain correct disentangled features. It also uses a consistency loss and a perceptual loss to enhance the disentanglement.

DRNET~\cite{denton2017unsupervised} adopts a $L_2$ loss as the reconstruction loss and uses a similarity loss together with an adversarial loss to ensure disentanglement. 
We ignore the task loss, because it is a two-stage scheme that the disentangled representations aim to be used in downstream tasks such as video prediction. 

InfoSwap~\cite{Gao_2021_CVPR} resorts to an information compression loss based on information bottleneck theory as the disentanglement loss. It uses an adversarial loss as the task loss of face identity swapping, and uses a consistency loss as the reconstruction loss.

MAP-IVR~\cite{liu2021activity} employs a cosine similarity loss to enforce orthogonality between the motion and appearance feature, in addition to the $L_2$ reconstruction loss which ensures the motion feature and the appearance feature capturing the dynamic and static information respectively. MAP-IVR also has no task loss since it is a two-stage scheme. The learned motion and appearance features to tackle the downstream task, e.g., activity image-to-video retrieval.

Besides, several works also introduce extra supervision to enforce disentanglement without explicit disentanglement loss function, such as DR-GAN~\cite{tran2017disentangled} and DNA-GAN~\cite{xiao2017dna}.

\subsubsection{Discriminative task}


Discriminative tasks sometimes also need reconstruction loss. For example, discriminative tasks can use a VAE backbone to do feature extraction, or leverage reconstruction loss to ensure the disentangled features can correctly recover the data. Discriminative tasks usually do not restrict any specific backbone models, 
they adopt the latent disentangled representation encoded by appropriate models 
such as VAE or GAN, based on which the task loss required by the target task
such as image classification, recommendation, neural architecture search etc., will be added.

For example, Hamaguchi et al.~\cite{Hamaguchi_2019_CVPR} add similarity loss and activation loss on the basis of using two pairs of VAEs to encode image pairs, which aims to make common features encode invariant factors in an input image pair. The two losses encourage the model to learn common features and specific features of images separately. It uses a classification loss as the task loss to achieve rare event detection, which also guides the training of the disentanglement encoders.


Wu et al.~\cite{wu2021vector} use an orthogonal loss to promote the independence between domain-invariant and domain-specific features. Meanwhile, they also use an adversarial mechanism to encourage the domain-specific features to capture more domain-specific information. They use a detection loss to do domain adaptive object detection.

Cheng et al.~\cite{cheng2021disentangled} use a gradient reverse layer and a class discriminative loss to minimize the class-specific information captured by the class-irrelevant encoder. They use a L1 reconstruction loss and a translation loss to ensure the disentangled features can correctly recover the data. They use a classification to accomplish few-shot image classification.

Table~\ref{tab: loss functions} summarizes designs of loss functions of these generative and discriminative tasks.

\section{Future Directions}
\label{sec:future}
Last but not least, we conclude this paper by pointing out some potential interesting directions that deserve future investigations.

\red{
\spara{Deriving Better Theoretical Foundations} Although DRL has been empirically shown to be effective, there
still lack strict  mathematical guarantees for whether particular representations can always be completely disentangled, or to which degree can the disentanglement be. Moreover, is it possible to discover any theoretical connections between disentanglement and generalization or robustness? Therefore, the following two points may be a good start for future investigation. 
\begin{enumerate}
\item Proposing solid mathematical frameworks. Establish a rigorous mathematical foundation that can define and measure disentanglement precisely. This includes understanding the conditions under which disentanglement is possible and beneficial. 
\item Connecting to generalization and robustness. Develop theories that explain how disentangled representations improve generalization across different tasks and domains and enhance robustness against adversarial attacks.
\end{enumerate}

\spara{Improving Evaluation Metrics} 
It is very important to create more objective, standard, comprehensive and universally accepted benchmarks and metrics for evaluating the quality and effectiveness of disentangled representations. This will help in comparing different approaches more fairly and systematically, pushing forward the future research on DRL.

\spara{Enhancing Explainability towards A New Level} Indeed one major motivation of DRL is to learn explainable representations of data, however, it is a pity that none of existing DRL approaches are able to fully explain the semantic meaning of latent representations in vector space. As such we believe that continuing to enhance the capability of explanation to next level is still of great importance for further investigations.

\begin{enumerate}
\item Human-understandable representations. The ultimate goal of DRL is expected to aim for representations that are not only disentangled but also understandable by humans, facilitating explainable AI. This includes aligning the learned factors with human intuition and semantic meanings. 
\item Interactive disentangled representation learning. It may also be interesting to explore interactive learning environments where humans can guide the disentanglement process, ensuring that the learned representations align with user-defined concepts of interest.
\end{enumerate}

\spara{Disentangling Foundation Models}
Foundation models such as \textit{chatGPT} and \textit{Stable Diffusion} are now prevailing in the research community, which are powerful in various downstream tasks. The strength of foundation models mainly comes from large-scale training data and billions of deep neural network parameters. We believe that foundation models may benefit from DRL for the following points.

\begin{enumerate}
\item Adaptation to specific tasks. For a specific downstream task, there usually exists redundancy in foundation models. Existing methods always use transfer learning based techniques such as fine-tuning to adapt foundation models to specific tasks. However, transfer learning can not distinguish which part of the knowledge in the pretrained model is relevant to the task, which may result in redundancy as well. DRL is able to disentangle the task-relevant parts and the task-irrelevant parts, having the potential to make the adaption more precise and efficient.
\item Interpretability of Foundation model. Foundation models are powerful and even have the reasoning capability to some extent, however, it is still a black box model that lacks the ability to explain. DRL has the potential to make foundation models more transparent and interpretable. For example, DRL might be able to identify the roles of representations within different network modules and decompose the inference process into human-understandable individual components.
\end{enumerate}

\spara{Exploring the Power of DRL in Advanced Real-world Scenes} 
As mentioned before, theoretical research of DRL mainly focuses on simple datasets, which might not keep the path with various advanced applications (e.g., visual question answering, text-to-image generation, and text-to-video generation) and models (e.g., Diffusion Models) which involve real-world datasets and scenes. We believe it is necessary to explore the power of DRL to facilitate these advanced real-world applications as follows.

\begin{enumerate}
    \item On the one hand, we may need to design more effective model architectures suited for real-world disentanglement that usually involve coarse-grained complex factors. For example, Capsule networks might be a promising direction. Traditional dimension-wise DRL such as VAE-based methods might not be suitable for real-world scenes.
    \item On the other hand, we should explore the power of existing ideas of DRL in various advanced tasks and models, e.g., diffusion text-to-image generation. Considering the factorability, robustness, and generalization brought by DRL might facilitate these tasks.
    \item In addition, the interdisciplinary research trending requires us to solve real-world problems across various domains such as healthcare, autonomous vehicles, and finance. This being the case, understanding domain-specific needs can inspire new techniques for DRL. 
\end{enumerate}

\spara{Paying Attention to Ethical and Fair AI} Last but not least, we should always keep in mind to address ethical concerns and biases in AI systems through leveraging disentangled representations to identify and mitigate sources of bias and unfairness as much as we can.

}

 {

 }
%



%

\begin{IEEEbiography}[{\includegraphics[width=0.8in,height=1.0in,clip,keepaspectratio]{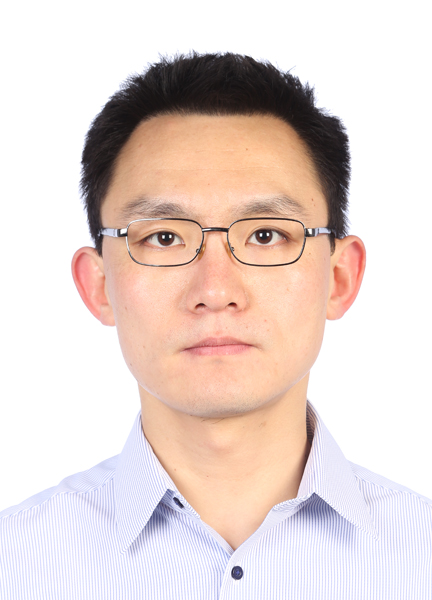}}]{Xin Wang}\scriptsize
 is currently an Associate Professor at the Department of Computer Science and Technology, Tsinghua University. He got both of his Ph.D. and B.E degrees in Computer Science and Technology from Zhejiang University, China. He also holds a Ph.D. degree in Computing Science from Simon Fraser University, Canada. His research interests include multimedia intelligence, machine learning and its applications. He has published over 150 high-quality research papers in ICML, NeurIPS, IEEE TPAMI, IEEE TKDE, ACM KDD, WWW, ACM SIGIR, ACM Multimedia etc., winning three best paper awards including ACM Multimedia Asia. He is the recipient of ACM China Rising Star Award, IEEE TCMC Rising Star Award and DAMO Academy Young Fellow.
\end{IEEEbiography}

\begin{IEEEbiography}[{\includegraphics[width=0.8in,height=1.0in,clip,keepaspectratio]{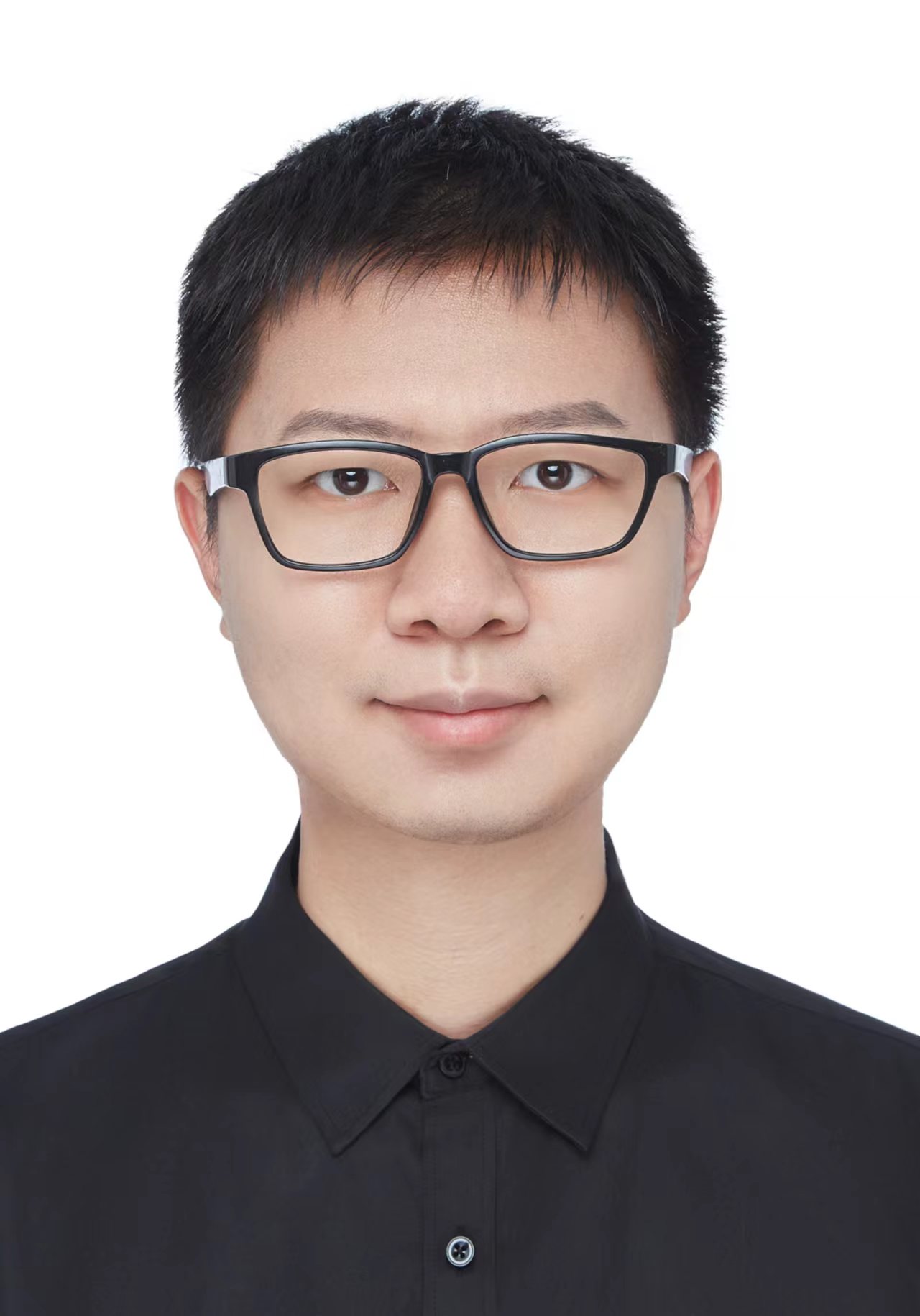}}]{Hong Chen} \scriptsize
received B.E. from the Department
of Electronic Engineering, Tsinghua University,
Beijing, China in 2020. He is currently a PH.D.
candidate in the Department of Computer Science and Technology of Tsinghua University. His
main research interests include auxiliary learning and multi-modal learning. He has published several papers in top conferences and journals including NeurIPS, ICML, IEEE TPAMI, etc. 
\end{IEEEbiography}


\begin{IEEEbiography}[{\includegraphics[width=0.8in,height=1.0in,clip,keepaspectratio]{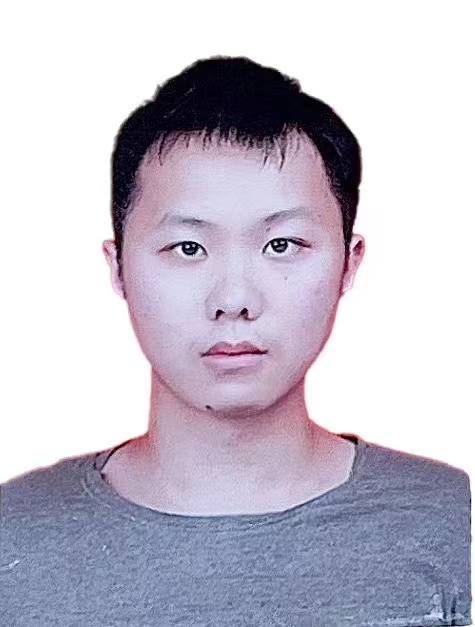}}]{Si'ao Tang}\scriptsize
 is a master candidate at Tsinghua-Berkeley Shenzhen Institute, Tsinghua University, majored in Data Science and Information Technology. His research interests include machine learning, multimedia intelligence, video understanding, etc.
\end{IEEEbiography}

\begin{IEEEbiography}[{\includegraphics[width=0.8in,height=1.0in,clip,keepaspectratio]{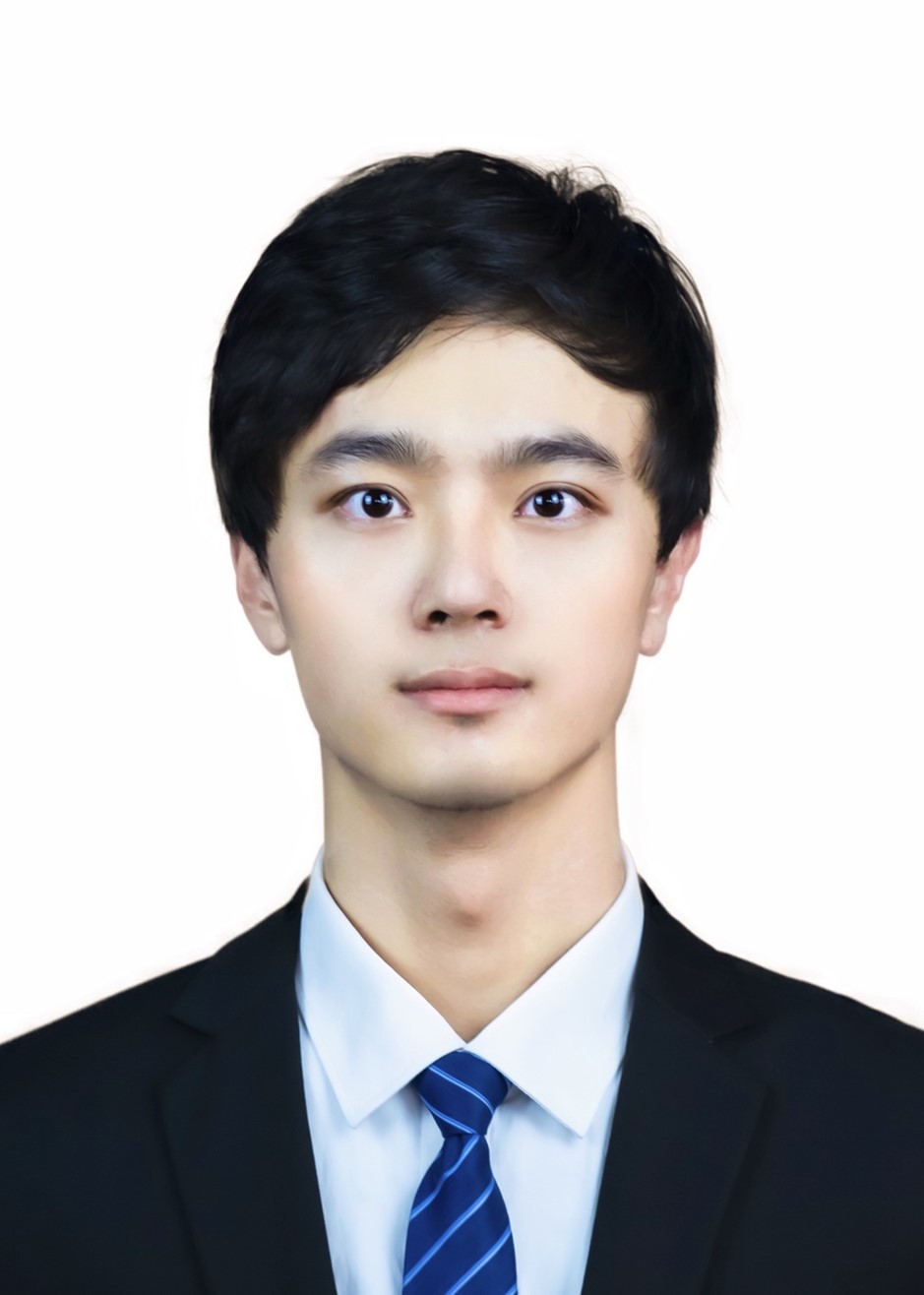}}]{Zihao Wu}\scriptsize
is currently working toward the master’s degree in computer science and technology with Tsinghua University, Beijing, China. He recieved his B.E. degree from the Department of Computer Science, Tongji University.
His research interests include machine learning, multimedia intelligence, and recommendation.
\end{IEEEbiography}

\begin{IEEEbiography}[{\includegraphics[width=0.8in,height=1.0in,clip,keepaspectratio]{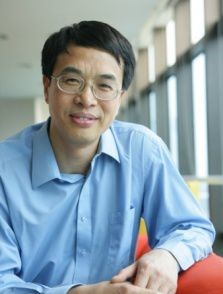}}]{Wenwu Zhu}\scriptsize
is currently a Professor in the Department of Computer Science and Technology at Tsinghua University. He also serves as the Vice Dean of National Research Center for Information Science and Technology, and the Vice Director of Tsinghua Center for Big Data. 
His research interests are in the area of data-driven multimedia networking and Cross-media big data computing. He has published over 380 referred papers and is the inventor or co-inventor of over 80 patents. He received eight Best Paper Awards, including ACM Multimedia 2012 and IEEE Transactions on Circuits and Systems for Video Technology in 2001 and 2019.  

He served as EiC for IEEE Transactions on Multimedia from 2017-2019. He served in the steering committee for IEEE Transactions on Multimedia (2015-2016) and IEEE Transactions on Mobile Computing (2007-2010), respectively.
He is an AAAS Fellow, IEEE Fellow, SPIE Fellow, and a member of The Academy of Europe (Academia Europaea).
\end{IEEEbiography}




\end{document}